%% file: main.tex
\pdfoutput=1

\documentclass[11pt]{article}

\usepackage[final]{acl}

\usepackage{times}
\usepackage{latexsym}

\usepackage[T1]{fontenc}

\usepackage[utf8]{inputenc}

\usepackage{microtype}

\usepackage{inconsolata}

\usepackage{graphicx}

\usepackage{booktabs} 
\usepackage{multirow} 
\usepackage{tabularx} 
\usepackage{subfigure}

\newtheorem{definition}{Definition}

\usepackage{stfloats}

\title{Re-TASK: Revisiting LLM Tasks from Capability, Skill, and Knowledge Perspectives}

\author{
  Zhihu Wang$^{1*\dag}$, Shiwan Zhao$^{2*}$, Yu Wang$^{3*}$, Heyuan Huang$^{1}$, Sitao Xie$^{4}$\\
  \textbf{Yubo Zhang}$^{1}$, \textbf{Jiaxin Shi}$^{1}$, \textbf{Zhixing Wang}$^{1}$, \textbf{Hongyan Li}$^{1}$, \textbf{Junchi Yan}$^{4}$  
  \\
  $^{1}$Huawei Technologies Ltd.\quad
  $^{2}$Nankai University\\
  $^{3}$Xi'an Jiaotong University\quad
  $^{4}$Shanghai Jiaotong University\\
  \texttt{\{wangzhihu3,huangheyuan,zhangyubo20,wangzhixing11\}@huawei.com} \\
  \texttt{zhaosw@gmail.com}\quad  \texttt{uyleewang@stu.xjtu.edu.cn}\quad
  \texttt{shijx12@gmail.com}\\ \texttt{lihongyan\_csu@163.com}\quad
  \texttt{\{xiest0518,yanjunchi\}@sjtu.edu.cn} 
}

\begin{document}
\maketitle
\renewcommand{\thefootnote}{}
\footnotetext{*\ Equal Contribution.}
\footnotetext{\dag\ Corresponding Author.}
\renewcommand{\thefootnote}{\arabic{footnote}}

\begin{abstract}
The Chain-of-Thought (CoT) paradigm has become a pivotal method for solving complex problems with large language models (LLMs). However, its application to domain-specific tasks remains challenging, as LLMs often fail to decompose tasks accurately or execute subtasks effectively. This paper introduces the \textbf{Re-TASK} framework, a novel theoretical model that \textbf{\underline{Re}}visits LLM \textbf{\underline{T}}asks from c\textbf{\underline{A}}pability, \textbf{\underline{S}}kill, and \textbf{\underline{K}}nowledge perspectives, drawing on the principles of Bloom's Taxonomy and Knowledge Space Theory. 
While CoT provides a workflow-centric perspective on tasks, Re-TASK introduces a Chain-of-Learning (CoL) paradigm that highlights task dependencies on specific capability items, further broken down into their constituent knowledge and skill components. To address CoT failures, we propose a Re-TASK prompting strategy, which strengthens task-relevant capabilities through targeted knowledge injection and skill adaptation. Experiments across diverse domains demonstrate the effectiveness of Re-TASK. In particular, we achieve improvements of 45.00\% on Yi-1.5-9B and 24.50\% on Llama3-Chinese-8B for legal tasks. These results highlight the potential of Re-TASK to significantly enhance LLM performance and its applicability in specialized domains. We release our code and data at \href{https://github.com/Uylee/Re-TASK}{https://github.com/Uylee/Re-TASK}.
\end{abstract}

\section{Introduction}

As the scale of large language models (LLMs) continues to increase, their general capabilities in natural language processing (NLP) tasks have shown substantial improvements. However, despite these advances, these models often struggle with complex reasoning tasks, particularly those that are domain-specific. The Chain-of-Thought (CoT) technique \citep{wei2022chain, wang2023selfconsistency, zhou2023leasttomost} has emerged as a promising paradigm by decomposing complex tasks into a series of subtasks in a divide-and-conquer manner. Yet, the application of CoT to domain-specific tasks faces significant challenges in both task decomposition \citep{kambhampati2024can} and subtask execution \citep{lightman2024lets}, due to a lack of domain knowledge and specialized capabilities.

The idea that an individual's capabilities directly influence task performance is well supported by educational theories, notably Bloom's Taxonomy \citep{bloom2010taxonomy} and Knowledge Space Theory (KST) \citep{doignon1985spaces}. Bloom’s Taxonomy outlines how educational objectives are achieved through structured instructional activities, each involving specific knowledge and cognitive processes. Similarly, KST emphasizes the sequential dependencies between learning items, forming \emph{“learning pathways”} that guide learners from foundational knowledge to mastery.

\begin{figure*}
\centering
\includegraphics[width=0.965\textwidth]{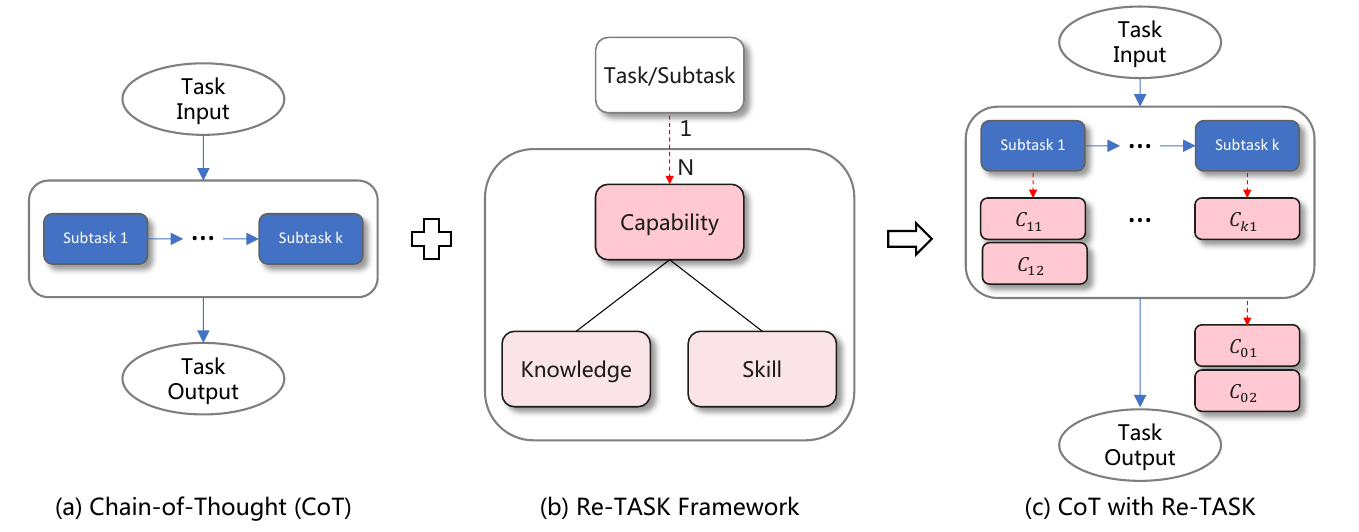}
\caption{The Chain-of-Thought (CoT) provides a workflow perspective on tasks (blue arrow), while the Re-TASK framework introduces a Chain-of-Learning view (red dashed arrow), demonstrating how tasks and subtasks depend on various $N$ capability items. Combining CoT with Re-TASK enhances CoT's performance in both task decomposition and subtask execution. Here, \(C_{ij}\) represents the capability item associated with subtask \(i\) (where \(i=1,\cdots,k\)), \(j\) denotes the capability item index, and \(C_{0j}\) refers to capability items associated with the overall task for task decomposition.}
\label{fig:framework}
\end{figure*}

Building on these insights into the Chain-of-Learning (CoL), we introduce the Re-TASK framework, which revisits LLM tasks through the lenses of capability, skill, and knowledge (see Figure~\ref{fig:framework}(b)). This framework posits that the successful completion of tasks\footnote{We use the terms “task” and “subtask” interchangeably, depending on the context.} depends on sequentially mastering multiple capability items, with each item further dissected into its constituent aspects of knowledge and skills. 

Our framework argues that failures in the CoT paradigm, particularly in task decomposition and subtask execution, stem from a lack of the necessary capabilities, due to either insufficient knowledge or inadequate knowledge-skill adaptation (abbreviated as skill adaptation). First, LLMs may lack relevant knowledge due to limited access to proprietary data or issues related to data timeliness. Second, even when knowledge is available, LLMs often struggle to apply it effectively because of inadequate knowledge-skill adaptation, resulting in suboptimal performance. While techniques such as retrieval-augmented generation \citep{lewis2020retrieval} can inject knowledge into the context, models may still fail if there is insufficient skill adaptation to effectively utilize the retrieved knowledge.

To address these issues, we propose integrating the CoT paradigm with the Re-TASK framework to improve LLM performance, as shown in Figure~\ref{fig:framework}. Specifically, we identify core capability items linked to the overall task and its corresponding subtasks, then strengthen these capabilities through targeted knowledge injection and skill adaptation using a deliberately designed prompting strategy, Re-TASK prompting (see Figure~\ref{fig:prompt-strategy}). The capability items represent demonstrations of knowledge-skill adaptation, such as conceptual knowledge understanding and procedural knowledge applying. Notably, knowledge itself can be treated as a special capability item, as in the case of knowledge injection through recalling or retrieving. We then adopt in-context learning (ICL) techniques \citep{brown2020language, dong2022survey} to enhance the corresponding capabilities by carefully arranging these demonstrations within the prompt.

We conduct extensive experiments with open-source LLMs to evaluate the effectiveness of the Re-TASK framework in enhancing CoT performance across various domain-specific tasks. Specifically, we included datasets from three major domains—finance, law, and STEM—comprising a total of five distinct datasets in both Chinese and English, along with corresponding multilingual models. This selection ensures that our results are both meaningful and broadly applicable. By incorporating relevant capability items that inject domain knowledge or improve skill adaptation, we observe significant improvements in task performance. Additionally, we extend our experiments to LLMs of different scales, demonstrating that while model capabilities generally improve with scale, the Re-TASK framework continues to enhance performance effectively. 

Our main contributions are summarized as follows:
\begin{itemize}
\item We introduce the Re-TASK framework, a novel theoretical model that revisits LLM tasks from the perspectives of capability, skill, and knowledge, offering a Chain-of-Learning view of tasks.
\item Our research reveals that many failures of the CoT approach in addressing domain-specific tasks stem from insufficient knowledge or inadequate skill adaptation.
\item We propose the Re-TASK prompting strategy, which integrates CoT with the Re-TASK framework to enhance LLM performance using in-context learning techniques.
\item Extensive experimental results demonstrate the effectiveness of the Re-TASK framework. 
\end{itemize}

\section{Related Work}
\label{related_work}

\subsection{Educational Theories}
Bloom's Taxonomy \citep{bloom2010taxonomy} provides a foundational framework that links learning objectives with instructional activities. It posits that achieving learning goals requires the completion of multiple interconnected activities, categorized along two key dimensions: knowledge and cognitive processes. The knowledge dimension outlines four types of knowledge: factual, conceptual, procedural, and metacognitive. The cognitive process dimension establishes a hierarchy of cognitive skills, encompassing six levels—remember, understand, apply, analyze, evaluate, and create—each linked to specific cognitive processes, totaling 19 distinct actions. By integrating these dimensions, Bloom's Taxonomy serves as a comprehensive guide for educators to design curricula and instructional strategies that foster deeper understanding and encourage higher-order thinking in students.

Knowledge Space Theory (KST) \citep{doignon1985spaces,falmagne2013knowledge,cosyn2021practical} offers a mathematical framework for modeling and assessing learners' knowledge within a specific domain. It identifies various knowledge states, defined as sets of problems or concepts that a learner can successfully solve or understand. The entirety of these possible knowledge states forms the knowledge structure, which delineates the relationships among different states. Learning pathways are the routes learners can take to transition from one knowledge state to another. By utilizing KST, educators can design effective educational interventions and personalized learning experiences, optimizing the learning process for each individual learner.

Building on these foundational educational theories, we propose the Re-TASK framework, which elucidates the dependence of tasks on various capability items. Each capability item is further broken down into its constituent aspects of knowledge and skills, highlighting the intricate relationships that contribute to task performance.

\subsection{Knowledge and Skills in LLMs}
Several studies have explored LLMs from the perspectives of knowledge and skill. KoLA \citep{yu2023kola} emphasizes the importance of world knowledge for LLMs and establishes a knowledge-oriented evaluation benchmark. In its approach to ability modeling, KoLA simplifies and selects from Bloom's learning theories to form four levels of knowledge-capability assessment: knowledge memorization, knowledge understanding, knowledge applying, and knowledge creating. Skill-it \citep{chen2024skill} posits that language models naturally acquire a sequence of skills from training data and formalizes the notion of a skill and an ordered set of skills in terms of associated data, differentiating this approach from traditional curriculum learning \citep{bengio2009curriculum}, which focuses on training models using progressively difficult examples. In continual pre-training experiments, Skill-it’s ordered learning of skills achieves faster convergence of validation loss compared to random sampling. RA-DIT \citep{lin2023ra} introduces a lightweight fine-tuning methodology that improves retrieval-augmented language models by enhancing both the relevance of retrieved knowledge and its effective utilization, marking a specialized form of knowledge and skill enhancement. MMLU \citep{hendrycks2020measuring} serves as a benchmark designed to measure the possession of world knowledge and problem-solving abilities. Reflecting on these developments, \citet{bengioBlog} have emphasized the integration of the world model and the inference machine in current LLMs. They suggest that to reason effectively, a robust world knowledge model and a powerful inference machine are necessary, advocating for their separation and simultaneous development to enhance reasoning capabilities.

\subsection{Prompting Strategies}
Various prompting strategies have been proposed to enhance model performance in solving complex, domain-specific tasks. One prominent approach is Chain-of-Thought (CoT) \citep{wei2022chain, wang2023selfconsistency, zhou2023leasttomost}, which decomposes a complex task into simpler subtasks, utilizing a divide-and-conquer strategy. Another notable approach involves integrating LLMs with knowledge. Some methods achieve this by retrieving knowledge from external sources, such as Retrieval-Augmented Generation (RAG) \citep{lewis2020retrieval, RAGSurvey}, while others enable the models to generate knowledge internally, rather than relying on retrieval \citep{liu2021generated}.
In-Context Learning (ICL) techniques \citep{brown2020language, dong2022survey} represent a significant advancement in prompting; ICL utilizes examples within the prompt itself, enabling the model to learn from context without the need for explicit fine-tuning.

In contrast, we propose Re-TASK prompting, which integrates CoT prompting with the Re-TASK framework. This approach leverages ICL techniques to enhance corresponding capabilities by carefully arranging demonstrations of capability items within the prompt, ultimately leading to improved overall task performance.

\section{Re-TASK}
\label{approach}

\subsection{Re-TASK Framework}
We begin by defining several key concepts within our framework: tasks, capability items, knowledge, and skills. We then elucidate how these elements interconnect to establish the structured Re-TASK framework, as illustrated in Figure~\ref{fig:framework}(b). These concepts parallel Bloom's Taxonomy, where educational objectives—comparable to tasks in our framework—are systematically achieved through structured instructional activities, analogous to capability items. Each instructional activity involves the acquisition of specific types of knowledge and engages corresponding distinct cognitive processes, thereby facilitating effective knowledge-skill adaptation. The successful completion of an educational objective, or task, depends on mastering various capability items, each developed through these instructional activities. Knowledge Space Theory (KST) further reinforces this structured approach by highlighting sequential dependencies among capability items.

\begin{definition}
\textbf{(Task)} A task $\mathbf{T}$ is defined as a specific objective that LLMs are designed to achieve, characterized by a mapping from input $x$ to output $y$, facilitated by a task instruction $I$ and an optional context $ctx$. Formally, this relationship is expressed as $\mathbf{T}(ctx; I; x) = y$, where the semicolon denotes concatenation of inputs.
\end{definition}
The optional context $ctx$ can be leveraged for both knowledge injection or skill adaptation. By identifying corresponding capability items and incorporating a list of capability item demonstrations into $ctx$, this method aligns closely with typical in-context learning. In the Chain-of-Thought (CoT) paradigm, a task can be decomposed into a series of subtasks.

\begin{definition}
\textbf{(Knowledge)} A knowledge point $\mathbf{K}$ is defined as a text segment containing domain-specific knowledge that is essential for the performance of a task/subtask. In the context of LLMs, the knowledge $\mathbf{K}$ can also be implicit knowledge encoded within the model's parameters.
\end{definition}
According to Bloom's Taxonomy, we consider three types of domain knowledge: factual, conceptual, and procedural\footnote{Metacognitive knowledge is beyond the scope of LLMs and is not our primary focus.}. Each type plays a distinct role in task execution, contributing differently to the LLM’s ability to process and respond to task-specific demands.

\begin{definition}
\textbf{(Skill)} A skill $\mathbf{S}$ corresponds to the cognitive processes in Bloom's Taxonomy and is developed through related instructional activities, including knowledge recalling/retrieving, understanding, applying, and others.
\end{definition}

\begin{definition}
\textbf{(Capability Item)} A capability item \(\mathbf{C}\), corresponding to the concept of instructional activities in Bloom's Taxonomy, is a specific exercise or demonstration designed to guide LLMs in applying a particular skill \(\mathbf{S}\) to the relevant knowledge \(\mathbf{K}\), thereby facilitating knowledge-skill adaptation.
\end{definition}
Successfully completing a task \(\mathbf{T}\) requires the sequential mastery of multiple capability items. These items can be conceptualized as a chain of learning, where explicit dependencies among them are clearly defined. Figure~\ref{fig:framework}(c) illustrates these dependencies. A task generally involves comprehensive overall procedural knowledge (\(C_{01}\)), with its resolution corresponding to a capability item (\(C_{02}\)) that applies this knowledge in a manner akin to a CoT process. This procedural knowledge is further segmented into subtasks, each linked to specific knowledge and involving different capability items \(C_{ij}\). 

Note that the knowledge \(\mathbf{K}\) can be treated as a special capability item with the default skill of knowledge recalling or retrieving. Consequently, the task and its subtasks depend on two types of capability items: knowledge recalling (the knowledge itself) and knowledge-skill adaptation (e.g., knowledge understanding and applying).
For example, a sentencing prediction task in law can be decomposed into two subtasks: the first involves assessing the severity of the victim's injuries, while the second determines the sentence duration based on this severity. For the first subtask, the capability item \(C_{11}\) could represent knowledge of guidelines for assessing severity (\textit{knowledge recalling}), while \(C_{12}\) could demonstrate the application of \(C_{11}\) to assess the severity of the victim's injuries (\textit{knowledge-skill adaptation}), formatted as \textit{\{Instruction; Input: case description; Output: severity\}}.

\subsection{Capability Item Construction}
\label{capability_item}
\begin{figure*}[]
\centering
\includegraphics[width=\linewidth]{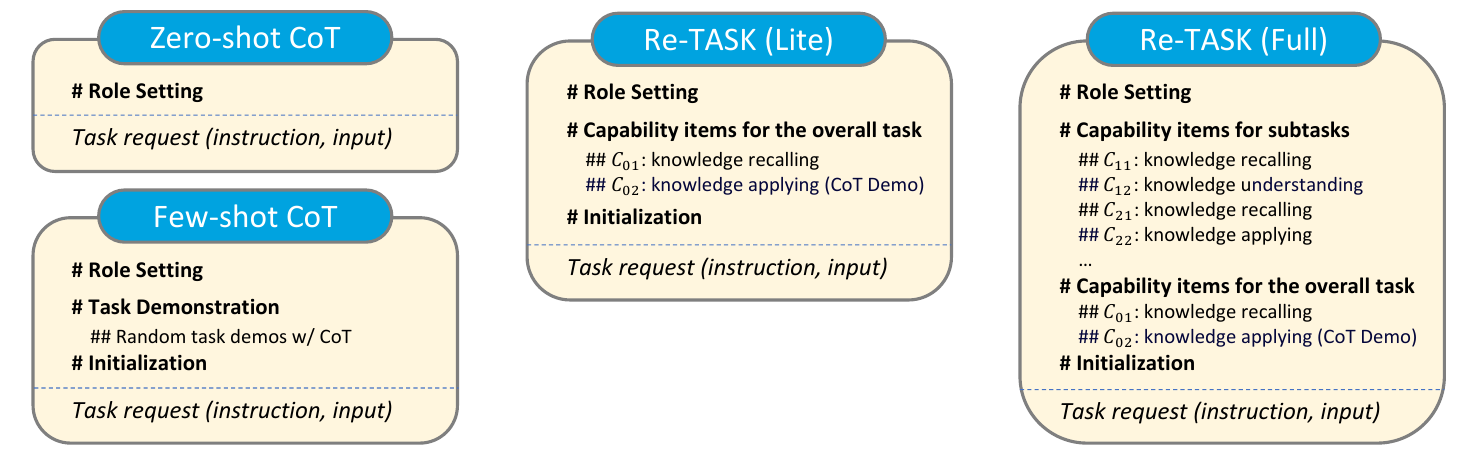}
\caption{Comparison of prompting strategies: Zero-shot CoT, Few-shot CoT, Re-TASK (Lite) prompting with only capability items for the overall task, and Re-TASK (Full) prompting incorporating all available capability items. In Re-TASK prompting, a task or subtask may be associated with any number of capability items. Note that other strategies, such as Self-Consistency (SC), are excluded for brevity.}
\label{fig:prompt-strategy}
\end{figure*} 
To effectively enhance the performance of LLM tasks, we identify key capability items across three categories: 1) recalling relevant knowledge, 2) understanding conceptual knowledge, and 3) applying procedural knowledge. The first category pertains to knowledge itself, while the latter two focus on knowledge-skill adaptation. Below, we present exemplary capability items for each category:

1) Knowledge retrieval: This involves identifying the relevant knowledge points for a given task or subtask and retrieving them from external sources. It may also include recalling internal knowledge points stored within the LLM.

2) Instances of conceptual knowledge: This involves providing examples that illustrate conceptual knowledge, helping to deepen and reinforce understanding.

3) Execution of procedural knowledge: This capability is crucial for tasks that involve following ordered steps or procedures, such as technical troubleshooting, recipe preparation, or complex calculations. It represents the practical application of procedural knowledge.

Notably, identifying the capability items associated with a task is a relatively straightforward process. For each given task, we begin by determining the relevant knowledge points, followed by identifying the corresponding skills required for effective task resolution, either understanding or applying.

\subsection{Re-TASK Prompting}

By combining CoT with the Re-TASK framework, we can identify core capability items linked to the overall task and its corresponding subtasks, subsequently strengthening these capabilities to enhance subtask performance and, ultimately, overall task performance. Specifically, we enhance the capability items through targeted knowledge injection and skill adaptation using a deliberately designed prompting strategy known as Re-TASK prompting, as illustrated in Figure~\ref{fig:prompt-strategy}. The capability items serve as demonstrations of knowledge-skill adaptation, including knowledge recalling/retrieving, knowledge understanding, and knowledge applying. 

We carefully arrange the demonstrations within the prompt according to their dependencies, following the chain of learning. Specifically, we sequence the capability items \(C_{ij}\) for each subtask \(i\). If multiple items are associated with the same subtask, we prioritize the knowledge itself (i.e., the capability item of knowledge retrieval) first, followed by more advanced items such as knowledge understanding and applying. Finally, we include the overall procedural knowledge \(C_{01}\) along with its applying \(C_{02}\) at the end of the prompt.

\section{Experimental Setup}

\subsection{Tasks and Datasets}
We selected the sentencing prediction task in the law domain, the financial course examination task in the finance domain, and multiple-choice question tasks in three STEM domains—mathematics, biology, and physics—to validate the effectiveness of the Re-TASK framework.

The sentencing prediction task in the law domain involves predicting the appropriate sentencing range for criminal offenders based on the factual descriptions in criminal cases. This process requires a comprehensive understanding of legal statutes and sentencing principles, making it a suitable challenge for validating the Re-TASK framework. 
For dataset construction, we utilized the publicly available CAIL dataset (China AI Law Challenge) \citep{xiao2018cail2018}, sampling 200 test instances. More details are presented in the Appendix~\ref{appendix:legaldataset}.

In the finance domain, we selected the financial course examination task in the FinanceIQ dataset \cite{zhang2023xuanyuan20largechinese}. The FinanceIQ dataset assesses LLMs' knowledge and reasoning abilities in financial contexts, evaluating their grasp of domain-specific knowledge. The financial dataset consists of multiple-choice questions, and the test set for the FinanceIQ dataset contains 178 instances. 

To evaluate the reasoning capabilities in STEM fields, we conducted experiments on the MMLU-Mathematics, Biology, and Physics benchmarks \citep{hendrycks2020measuring}. The test sets for these benchmarks contain 276, 144, and 102 multiple-choice questions, respectively. During dataset construction, a few mathematics questions entered an infinite loop while generating capability items. As a result, we removed these instances from the mathematics dataset. Further details can be found in the Appendix~\ref{appendix:datasets_fin_math}.

\subsection{Settings}

Given that both the legal and financial datasets are in Chinese, we opted for popular Chinese LLMs. Specifically, we selected the chat versions of Qwen1.5 \citep{bai2023qwen}, Llama3-Chinese \citep{ChineseLLaMAAlpaca3}, and Yi-1.5 \citep{young2024yi} for validation. For the mathematics, biology, and physics datasets, which are in English, we employed a different set of LLMs, including Llama3 \citep{dubey2024llama3herdmodels}, Mistral \citep{jiang2023mistral}, and Qwen1.5, to validate our results. To further verify the scalability of our framework, we conducted experiments with LLMs of varying scales, using the Qwen1.5 series with 7B, 14B, and 32B parameters on the legal dataset. More setting details are offered in the Appendix~\ref{sec:exp_details}.

We developed Re-TASK prompting strategies that integrate demonstrations of capability items to validate their effectiveness in improving the performance of CoT on domain-specific tasks. Our baselines include Zero-shot CoT, Few-shot CoT, Plan-and-Solve~\cite{wang2023plan}, STEP-BACK~\cite{zheng2023take}, and a knowledge-injected CoT variant, referred to as Re-TASK (+K0). In Re-TASK (+K0), ground-truth knowledge relevant to each instance is directly provided as context to guide the model’s reasoning process; therefore, this setting can be viewed as a variant of retrieval-augmented generation (RAG). We evaluated the performance of Few-shot CoT under two settings: One-shot and Three-shot learning, to align with the Re-TASK (Lite) and Re-TASK (Full) strategies, respectively. Additionally, we evaluated the performance of Zero-shot CoT with self-consistency. The prompt templates are illustrated in Figure~\ref{fig:prompt-strategy}. Details are shown in the Appendix~\ref{appendix:prompts}.

\subsection{Construction of Capability Items}
We utilized LLMs to assist in decomposing tasks and generating the capability items involved in Re-TASK. 
First, we predefined the capability types for each task based on the definition in Section~\ref{capability_item} and then leveraged larger LLMs, which are presumed to possess sufficient domain knowledge, to automatically generate these items.

We began by employing LLMs to decompose the tasks, which resulted in the identification of the overall procedural knowledge (\(C_{01}\)). Next, we instructed the LLMs to create a CoT demonstration using the generated knowledge as a knowledge applying capability item (\(C_{02}\)). Task decomposition also guides the creation of capability items for each subtask. For each subtask $i$, we instructed the LLMs to generate relevant conceptual or procedural knowledge ($C_{i1}$). For conceptual knowledge, we requested illustrative examples to enhance understanding, while for procedural knowledge, we sought CoT demonstrations to illustrate effective applying ($C_{i2}$). 

For financial tasks, we followed the complete procedure to generate capability items for both the overall task and its corresponding subtasks. In contrast, for other tasks, where the subtasks are relatively straightforward and do not require complex knowledge, we only generated capability items for the overall tasks.

\input{Tables/legal_results_new}

\section{Experimental Results}
We conducted extensive experiments across the law, finance, mathematics, biology, and physics domains to validate the effectiveness of Re-TASK. 

\subsection{Law Domain}

The results of the sentencing prediction task are presented in Table~\ref{tab:legal-results}. Notably, Re-TASK (Lite) achieves the best performance across all settings, surpassing Zero-shot CoT by an average of 27.17\%. Re-TASK (Lite) also outperforms the self-consistency version of Zero-shot CoT by a substantial margin of 26.84\% on average. Compared to Re-TASK (K0), which only incorporates the knowledge, Re-TASK (Lite) improves performance by 15.67\% on average. Specifically, Re-TASK (Lite) demonstrates a remarkable improvement of 45.00\% with the Yi-1.5-9B model. Furthermore, when the knowledge item (\(C_{01}\)) is excluded from the demonstration count, Re-TASK (Lite), with a single demonstration, shows significant gains compared to the One-shot CoT strategy. Step-Back also achieves a notable improvement, outperforming Zero-shot CoT by 18.00\%, though it still lags behind Re-TASK (Lite) by 9.17\%. These results underscore the effectiveness of the Lite version of Re-TASK prompting in the law domain.

\input{Tables/legal_efficiency}
We further compared the token lengths generated by different prompt strategies, including both the prompt and the completion, to assess efficiency. As shown in Table~\ref{table:tokens}, the inclusion of demonstrations generally resulted in an increase in token length. Specifically, the token length of Re-TASK (Lite) was comparable to that of One-shot CoT and shorter than that of Three-shot CoT.

\subsubsection{Model Scaling}
\begin{figure}[]
\centering
\includegraphics[width=0.5\textwidth]{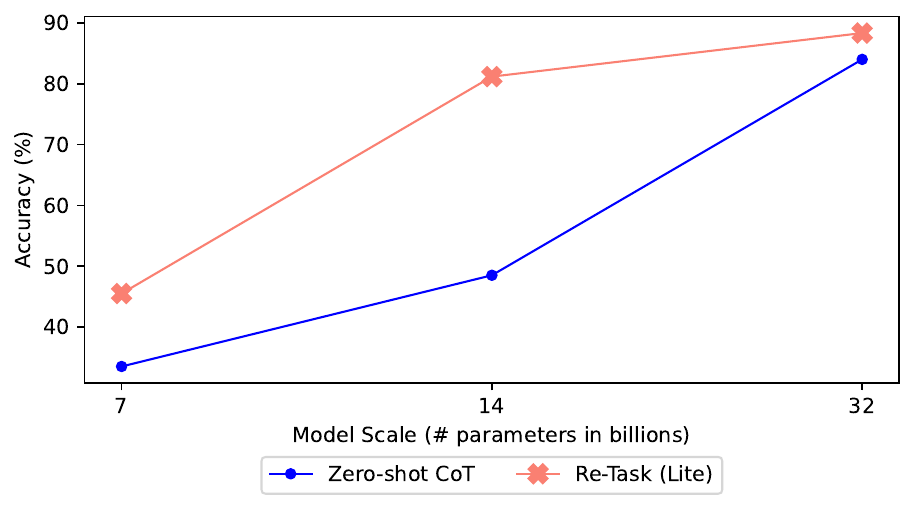}
\caption{Performance comparison of zero-shot CoT and Re-TASK (Lite) across different scales of Qwen1.5 models (7B, 14B, and 32B) on the sentencing prediction task.}
\label{fig:model-scale}
\end{figure}

To investigate the impact of model scales, we conducted experiments on the Qwen1.5 series across different scales. The results, presented in Figure~\ref{fig:model-scale}, reveal a consistent trend: as model size increases, the performance of both prompting strategies improves consistently. Notably, the inclusion of capability items led to significant and consistent gains in task performance. These findings underscore the limitations of LLMs in handling domain-specific tasks and demonstrate that our Re-TASK framework delivers substantial benefits even as model size scales.

\input{Tables/financial_results_new}
\subsection{Financial Domain}
\input{Tables/STEM}
As shown in Table~\ref{tab:financialresults}, Re-TASK exhibits substantial performance gains across all models in the FinanceIQ task. 
Baseline methods without demonstrations, including Zero-shot CoT, Zero-shot CoT with self-consistency, Step-Back, Plan-and-Solve, and Re-TASK (K0) strategies, show minimal improvements. In contrast, Re-TASK (Lite) and Re-TASK (Full) achieve notable average improvements of 5.06\% and 14.61\% over Zero-shot CoT, respectively. Specifically, Re-TASK (Lite) demonstrates an improvement of 19.67\% with the Yi-1.5-9B model and 16.29\% with the Llama3-Chinese-8B model. Compared to Re-TASK (K0), Re-TASK (Lite) achieves a 3\% improvement, demonstrating the importance of adding overall task capability items. Furthermore, under comparable numbers of demonstrations, Re-TASK (Lite) outperforms the One-shot CoT strategy by 1.46\%, while Re-TASK (Full) surpasses the Three-shot CoT strategy by 11.31\%, highlighting the effectiveness of our added capability item components. Collectively, these results underscore the pivotal role and remarkable efficacy of Re-TASK in improving task performance.

\subsection{STEM Domains}

The results for MMLU-Math, MMLU-Biology, and MMLU-Physics are presented in Table~\ref{tab:stem}. 
Re-TASK (Lite) demonstrates significant performance gains across all three datasets, particularly in the biology domain, where it outperforms Zero-shot CoT by an average of 18.29\%. In contrast, the self-consistency and plan-and-solve methods show inconsistent performance, with occasional improvements but no consistent gains overall. Moreover, Step-Back performs worse across the three datasets, likely due to suboptimal principles generated by the smaller-scale models themselves. When comparing with the same number of added demonstrations, Re-TASK (Lite) outperforms One-shot CoT across all three datasets. Notably, Re-TASK (Lite) achieves a significant 13.20\% improvement in the biology dataset, highlighting its strong and consistent performance in STEM tasks.

Compared to the results in the law domain, the improvements in the STEM datasets are relatively modest. This may be attributed to the fact that the capability items automatically generated by LLMs, which, while effective, may not be fully optimized. Nonetheless, our methods show that leveraging knowledge and capability items generated from larger models can be used to augment the performance of smaller models. Since our primary objective is to demonstrate the potential of our approach rather than to optimize the identification of capability items, these results still underscore the value of our method.

\section{Conclusion}\label{sec:conclusion}
In this paper, we introduced the Re-TASK framework, a novel approach that revisits LLM tasks through the lenses of capability, skill, and knowledge, aiming to address the limitations of the Chain-of-Thought (CoT) paradigm in complex, domain-specific tasks. By integrating Re-TASK with CoT, we systematically enhanced the performance of LLMs through targeted knowledge injection and skill adaptation, using a structured prompting strategy, Re-TASK prompting. Our extensive experiments across the law, finance, and STEM domains demonstrated that the Re-TASK framework significantly improves task performance, achieving substantial gains over baseline models and confirming the framework’s potential to enhance LLM capabilities across diverse domains. Our research opens new avenues for practitioners to deepen their understanding, evaluation, and improvement of LLMs.

\section*{Limitations}

While this study presents promising results, several limitations should be considered when interpreting the findings.

First, we rely on powerful LLMs to directly construct capability items for each task instance without incorporating the retrieval process. In a more practical scenario, an offline repository of capability items could be constructed based on a domain knowledge graph to support knowledge retrieval, along with corresponding capability items for knowledge understanding or applying. Retrieval-based methods (e.g., RAG) could then be employed to match capability items with tasks. This is part of our future work.

Second, although we used LLMs to automatically generate capability items, we did not focus on optimizing the generation process. As our primary objective was to demonstrate the potential of our approach rather than to refine the identification of capability items, we leave capability item optimization for future research.

Third, we did not thoroughly analyze the performance differences across various domains. These differences may stem from the extent to which LLMs possess domain-specific knowledge and their proficiency in applying corresponding skills, both of which influence the observed improvement ratios. A deeper investigation into these aspects would be a valuable direction for LLM diagnostics.

Finally, our study is limited to open-source models due to budget constraints, leaving the exploration of proprietary models to future work.

\bibliography{main}

\clearpage
\appendix

\section{All Experimental Results}

\subsection{Model Scaling}\label{appendix:model_scale}
The accuracy results for zero-shot CoT and Re-TASK (Lite) across various scales of Qwen1.5 models on the sentencing prediction dataset are shown in Table~\ref{Table:modelscale}, while the corresponding results on the FinancelQ dataset are presented in Table~\ref{Table:modelscale_fin}. Re-TASK demonstrates effectiveness across models of different sizes. A consistent trend emerges across different capability metrics: as the model scale increases, the performance of all prompting strategies improves accordingly. This demonstrates that Re-TASK is effective even on larger-scale models.
.
\input{Tables/legal_experimental_results_model_scale}
\input{Tables/model_scale_fin}

\subsection{Efficiency}
\input{Tables/financial_length}
\input{Tables/math_token}

To evaluate the efficiency of Re-TASK and eliminate the influence of computing platform performance on efficiency assessment, we adopt the sum of input and output tokens for LLMs as the metric of efficiency. Results for the financial and mathematical datasets are provided here. Table~\ref{tab:financial_length} and Table~\ref{tab:math_token} present the average token lengths for different methods on the FinanceIQ and MMLU-Math datasets across all questions. It can be observed that the token counts for Re-TASK are comparable to those of the corresponding One-shot and Three-shot CoT strategies. Thus, the efficiency of Re-TASK aligns closely with that of few-shot CoT strategy, demonstrating that Re-TASK introduces no additional overhead while achieving significantly higher accuracy.

\subsection{Case Study}\label{appendix:casestudy}

To better demonstrate the effectiveness of Re-TASK, we conduct a case study on the FinanceIQ and MMLU-Math datasets.

\begin{figure*}[ht]
\centering
\includegraphics[width=0.95\linewidth]{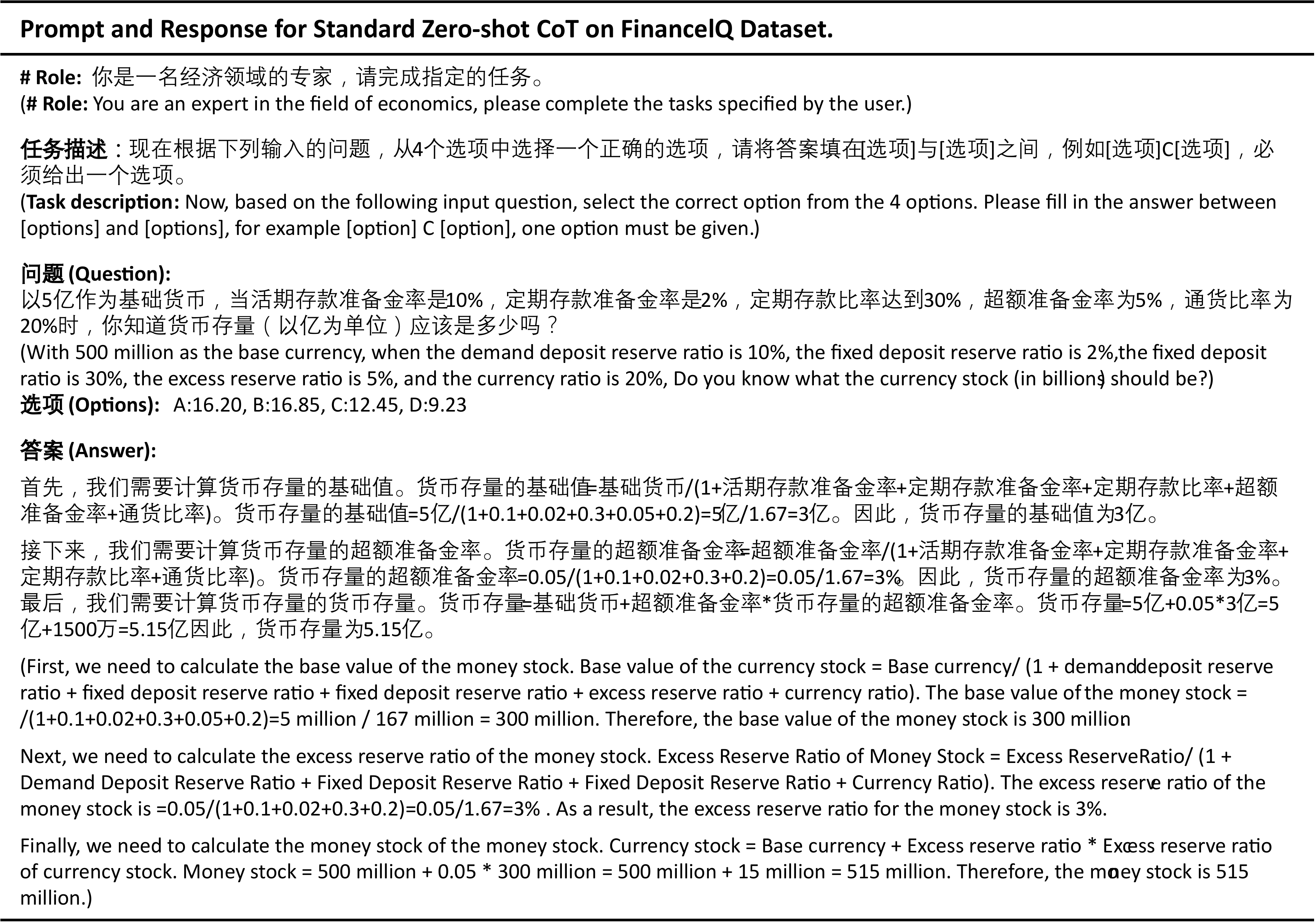}
\caption{An example of the Prompt and the Response for Zero-shot CoT strategy on FinanceIQ Dataset. }
\label{fig:financial_case_0cot}
\end{figure*} 

\begin{figure*}[ht]
\centering
\includegraphics[width=0.95\linewidth]{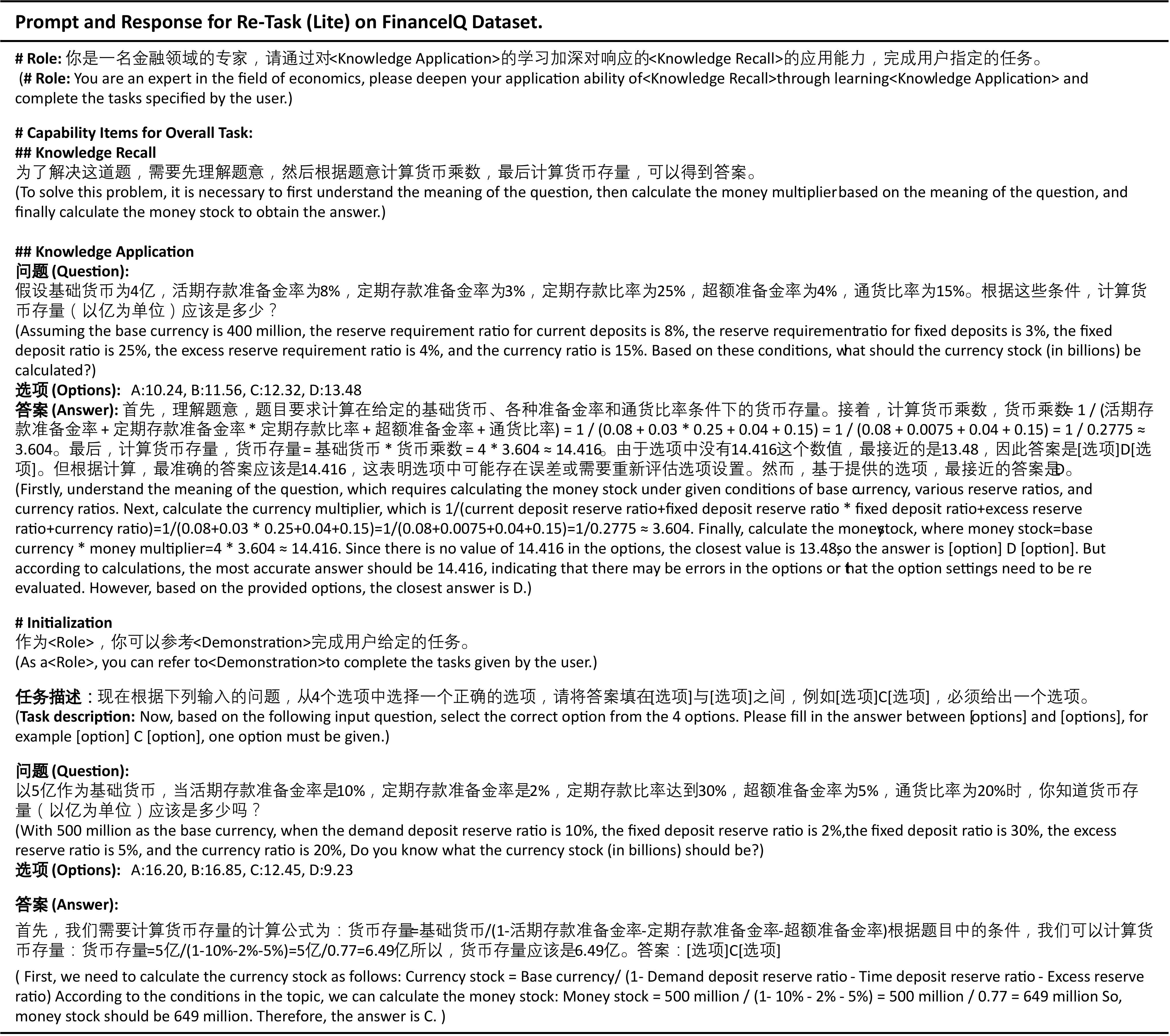}
\caption{An example of the Prompt and the Response for Re-TASK (Lite) strategy on FinanceIQ Dataset. }
\label{fig:financial_case_retask}
\end{figure*} 

\begin{figure*}[ht]
\centering
\includegraphics[width=\linewidth]{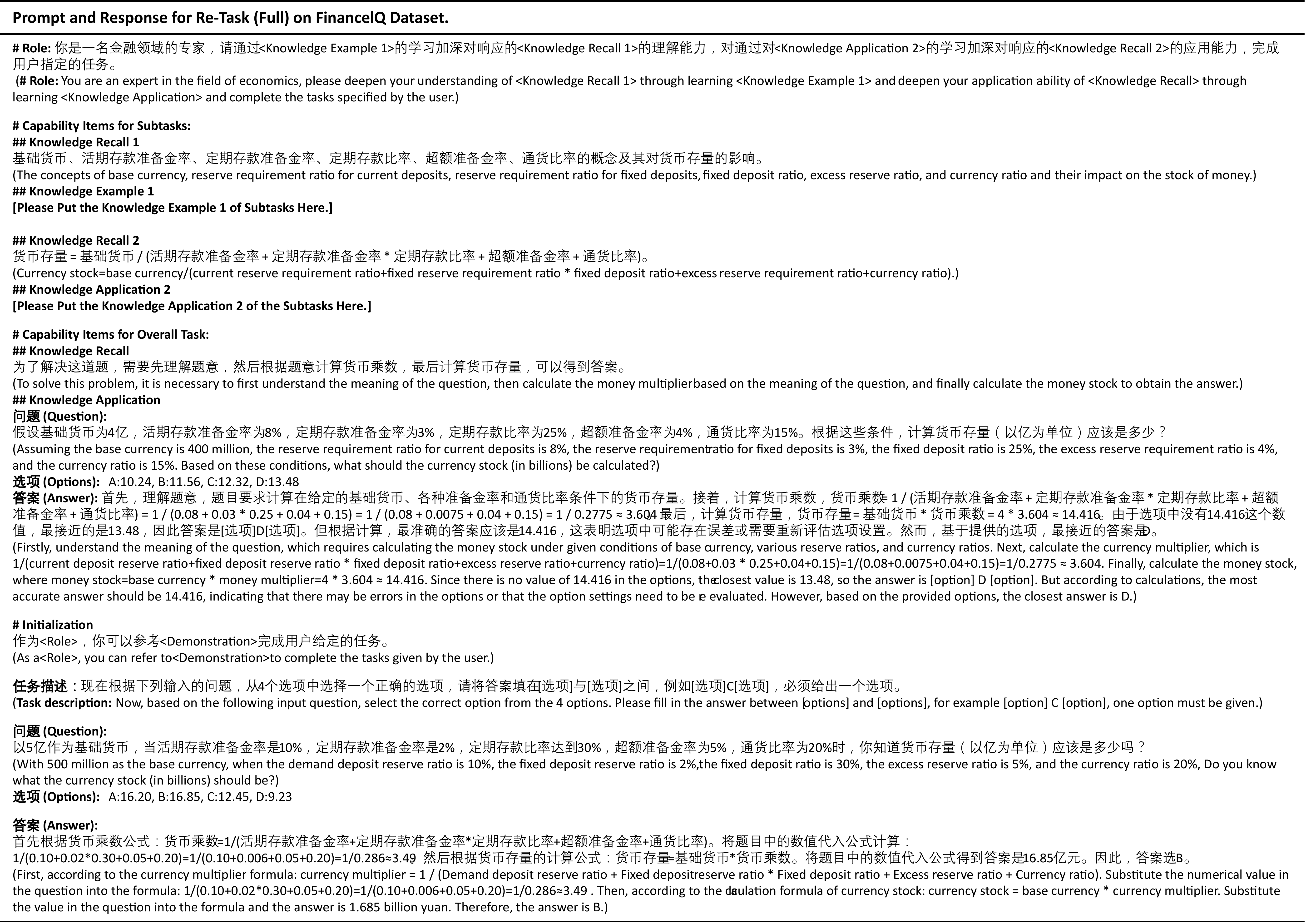}
\caption{An example of the Prompt and the Response for Re-TASK (Full) strategy on FinanceIQ Dataset. }
\label{fig:financial_case_retaskfull}
\end{figure*} 

Figures~\ref{fig:financial_case_0cot},~\ref{fig:financial_case_retask}, and~\ref{fig:financial_case_retaskfull} show the results of the same input-output example using three prompt strategies—Zero-shot CoT, Re-TASK (Lite), and Re-TASK (Full) —on the FinanceIQ dataset with the LLaMA3-8B-Chinese model. From the case, we can see that in the Zero-shot CoT strategy, the model fails to produce the correct answer because it lacks knowledge of the problem-solving steps and relevant knowledge about the topic, leading to repeated attempts and ultimately an incorrect result. In Re-TASK (Lite) strategy, while the model knows the correct formula for solving the problem, it makes errors during the numerical calculations, resulting in an incorrect answer as well. In Re-TASK (Full) strategy, the model applies the correct formula and arrives at the correct answer. This indicates that adding different capability items supplements the model’s knowledge and enhances its abilities in domain tasks, demonstrating the effectiveness of the Re-TASK framework.

\input{Tables/math_case1_base}
\input{Tables/math_case1_retask}

Table~\ref{tab:math_case1_base} and~\ref{tab:math_case1_retask} show one pair of comparison of Zero-shot CoT and Re-TASK (Lite) strategies with Llama3-8B on MMLU-Math dataset. For the Zero-shot CoT, it can be seen that though the LLM can stimulate the knowledge of the Pythagorean theorem, the following reasoning and calculation is still wrong. For the Re-TASK, we incorporate knowledge of the Pythagorean theorem through the prompt and demonstrated the applying of this knowledge through one capability item, and the correct reasoning and calculation result can be obtained. This comparison intuitively represents that capability items can effectively improve the reasoning.

\section{Experimental Settings}
\subsection{Models}\label{sec:models}
Seven models are utilized in this work: Qwen1.5-7B-Chat, Qwen1.5-14B-Chat, Qwen1.5-32B-Chat~\cite{bai2023qwen}, Yi-1.5-9B-Chat~\cite{young2024yi}, LLaMA-Chinese-8B-Instruct~\cite{ChineseLLaMAAlpaca3}, Llama3-8B~\cite{dubey2024llama3herdmodels}, and Mistral-7B~\cite{jiang2023mistral}. The Qwen1.5, Yi-1.5, Llama3, and Mistral-7B models are obtained as official chat versions from Hugging Face, whereas the LLaMA-Chinese-8B-Instruct model is sourced from Modelscope. In this study, Qwen2.5-72B is used for task decomposition and skill item generation on Chinese datasets (FinanceIQ and sentencing prediction tasks), while Llama3.1-70B is employed for task decomposition and skill item generation on the English dataset (MMLU). The model sources and licenses are shown in Table~\ref{tab:modelsource}.
\input{Tables/modelsource}

\subsection{Experimental Details}\label{sec:exp_details}

For the baseline prompts used in the experiments, we employed the standard zero-shot CoT and few-shot CoT templates, with the Step\_Back template and Plan-and-Solve template following the design from the original paper. The parameters for self-consistency were also based on the original work, and the prompt template used was consistent with the zero-shot CoT. Additionally, the selection of the normal one-shot and three-shot demonstrations was randomly chosen from the demonstration corpus.

For all experiments, except those involving Self-Consistency, the temperature is set to 0, and the top\_p parameter is set to 1. Because the temperature is set to 0, the results of multiple runs are consistent. For Self-Consistency, following the settings from the original paper~\cite{wang2022self}, the temperature is adjusted to 0.5, and top\_p is set to 0.5. Self-consistency involves the results of voting after running 20 rounds of zero-shot CoT.

\section{Experimental Datasets}

\input{Tables/datasource}
We used three publicly available datasets across five domains. The sources of the datasets used in this paper are shown in Table~\ref{tab:datasources}. The questions in these five datasets all require corresponding knowledge and skill adaptation to answer correctly, which is very suitable for verifying Re-TASK.

\subsection{Details of Sentencing Prediction Datasets}\label{appendix:legaldataset}
We construct the sentencing prediction dataset using data from the Cail2018 competition~\citep{xiao2018cail2018}, which is sourced from publicly available criminal legal documents on "China Judgments Online". Each record in the dataset includes case descriptions and factual details extracted from the legal documents. Additionally, each case provides the applicable legal articles, the charges against the defendant, and the length of the sentencing. The dataset contains approximately 2.68 million criminal law documents, covering 183 different charges and 202 legal articles, with sentences ranging from 0 to 25 years, life imprisonment, and the death penalty. Our focus is solely on the task of sentencing prediction.

The sentencing prediction task involves estimating the length of a defendant's sentence based on the descriptions and facts provided in the criminal legal documents. This task integrates five key elements from the dataset: the facts of the crime, the charges, the referenced legal articles, the defendant's name, and the length of the sentence.

Due to the large size and complexity of the original CAIL2018 dataset, we performed some preprocessing to better validate the retask. We implemented the following processing steps:

1) Several commonly encountered criminal law articles were selected to serve as the knowledge base. The original dataset was then filtered to include only criminal cases relevant to these articles. Additionally, we excluded data instances where the sentence prediction months were inherently implied based on the sentence range.

2) A suitable task instruction was designed, and the output format of the task was standardized.

3) The specific month lengths of sentences were converted into three broader sentencing categories: A (under 3 years), B (3 to 10 years), and C (over 10 years).

4) A series of robust and effective test prompt templates were designed for the task.

From the CAIL 2018 dataset, we constructed a 200-instance training set and a 200-instance test set. The training set and the test set are independent, with no overlapping instances. The distribution of the three sentencing categories (A, B, C) in both the training and test sets approximates a 1:1:1 ratio.

\begin{figure*}[h]
\centering
\includegraphics[width=\linewidth]{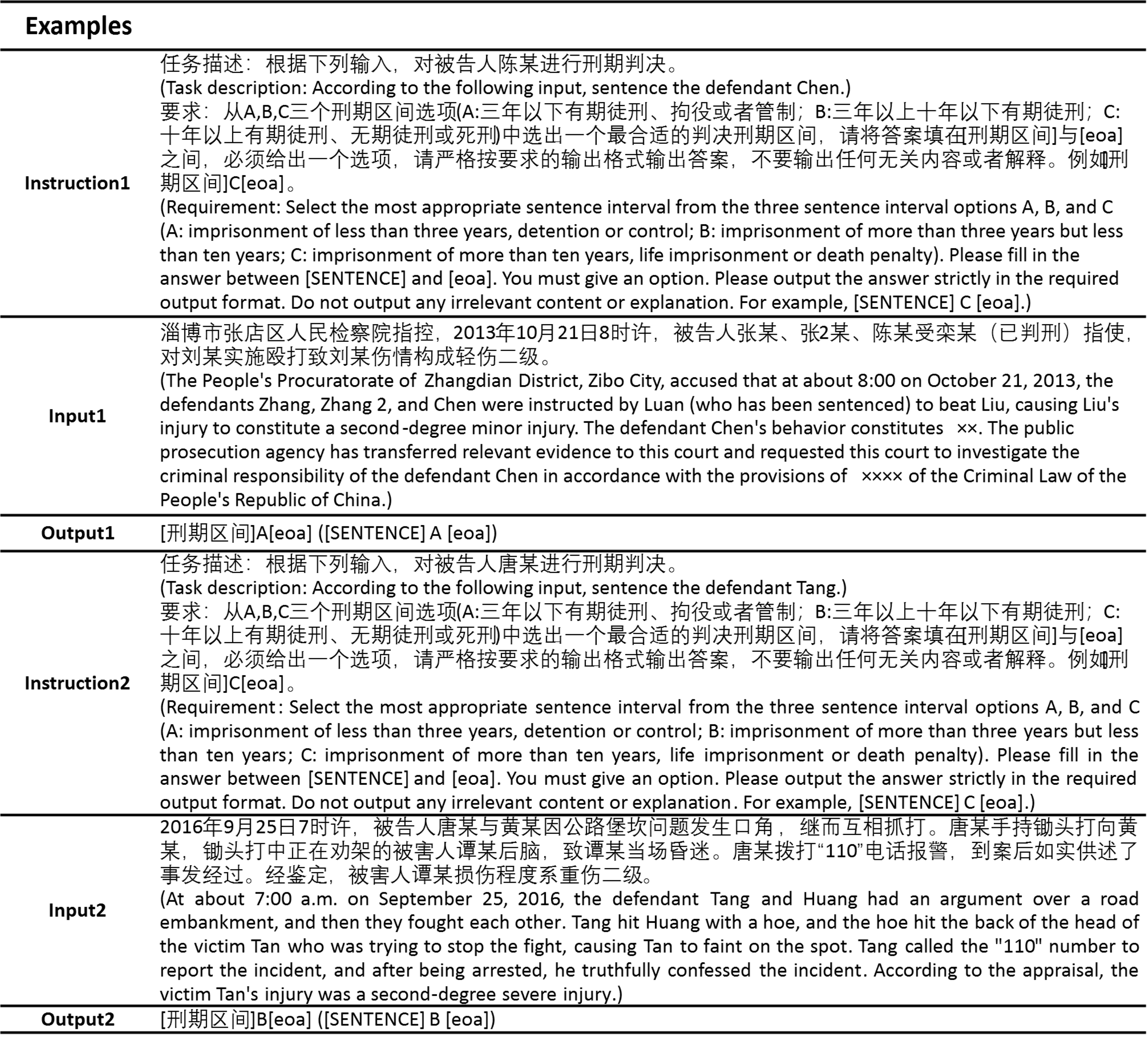}
\caption{Examples of the Sentence Prediction Task}
\label{fig:task}
\end{figure*}

Examples of sentence prediction tasks are illustrated in Figure~\ref{fig:task}.

We use Qwen2.5-72B to assist in the generation of capability items for the sentencing prediction task. Specifically, for a given instance, including the question, options, and answer, we first use LLMs to generate the knowledge behind the question. Then, based on the generated knowledge, we create the capability item. If the knowledge is conceptual, we generate an explanation or a demonstration example of the knowledge. If the knowledge is procedural, we generate a demonstration of how the knowledge is applied. More prompt templates are shown in the Appendix~\ref{appendix:prompts}.

\subsection{Details of MMLU datasets}

We use MMLU-high-school Mathematics, MMLU-college Biology and MMLU-college Physics datasets in this paper. For the MMLU-Math dataset, we utilized the original 304 questions. We used Llama3.1-70B to generate knowledge, but for some questions, the model entered into a feedback loop, rendering the generated knowledge invalid. In some cases, repeated attempts still resulted in the same feedback loop. As a result, we removed these problematic question-answer pairs, ultimately filtering out 276 questions. For the MMLU-Physics and MMLU-Biology datasets, such issues were rare, so we kept all instances and used the original datasets.

We use Llama3.1-70B to assist in the generation of capability items for the MMLU datasets. The generation method is consistent with the one used for the sentencing prediction task in Appendix~\ref{appendix:legaldataset}.

\subsection{Details of FinanceIQ datasets}\label{appendix:datasets_fin_math}

For the legal and MMLU datasets, since the tasks are relatively simple, we only constructed capability items related to the task, without building capability items for subtasks. When conducting processing for the FinanceIQ dataset, we needed to build both task-related capability items and subtask-related capability items, which made the process more complex. However, our analysis revealed that a significant portion of the FinanceIQ dataset consists of knowledge-intensive questions that can be addressed solely by providing factual knowledge, without requiring high levels of comprehension or reasoning abilities. These questions can be directly answered using RAG (retrieval-augmented generation), making it difficult to identify specific subtasks or construct subtask items. Consequently, we initially employed the Qwen2.5-72B model to assist in a filtering process, which effectively eliminated simpler questions that could be answered directly with factual knowledge.

During the generation process, we encountered special situations: Some questions resulted in infinite answer loops, particularly with the Llama series models, preventing them from generating knowledge or capability items properly. This issue could affect the subsequent validation of the Re-TASK framework. Due to the issues, we removed this portion of the data from the FinanceIQ dataset. Ultimately, we constructed our financial dataset.

We utilize Qwen2.5-72B to assist in the task decomposition of instances in the FinanceIQ dataset. In our experiments, we found that the subtasks in FinanceIQ still present a certain level of difficulty, requiring subtask-level knowledge and capability items. Therefore, for the financial dataset, we generated both task-related capability items and subtask-related capability items. We continue to use Qwen2.5-72B to help generate these capability items for FinanceIQ datasets. The generation process for task-related and subtask-related capability items is similar: we first provide the question, options, and answer to the LLM, allowing it to generate the knowledge; then, we feed the generated knowledge back to the model to generate the corresponding capability items.

\section{Experimental Prompts}\label{appendix:prompts}
\subsection{Law Domain}
\begin{figure*}[ht]
\centering
\includegraphics[width=0.95\linewidth]{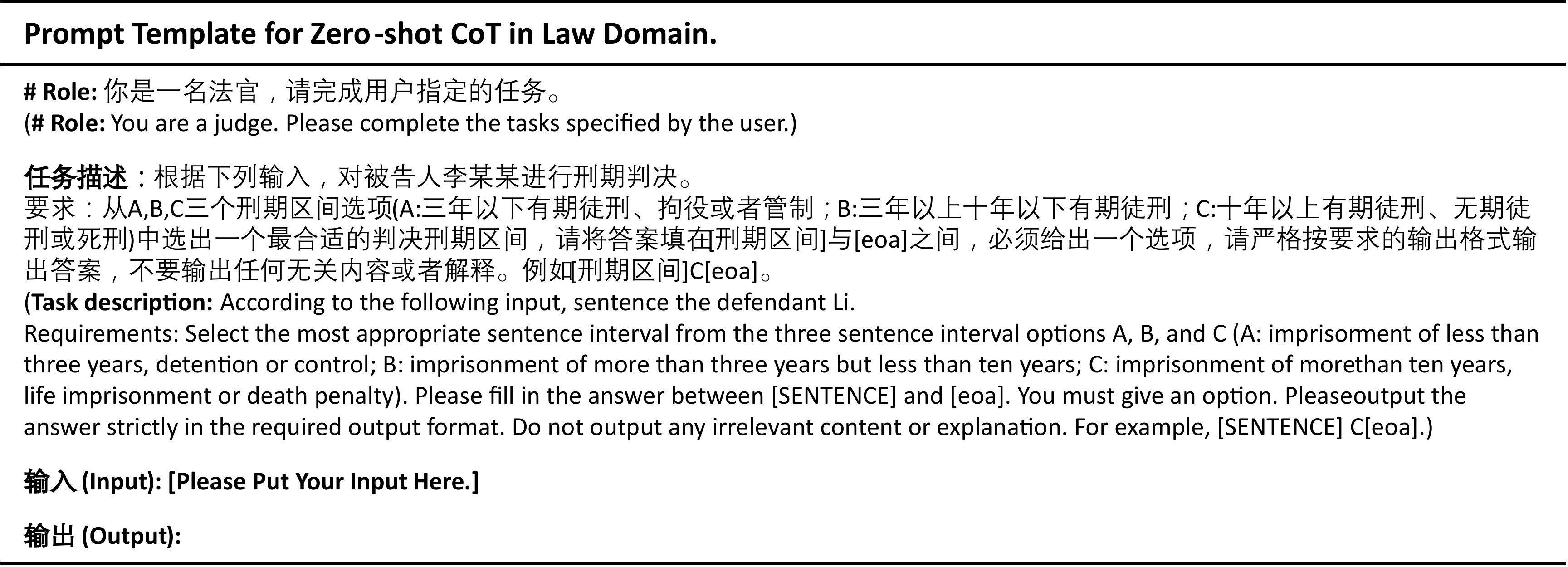}
\caption{The Prompt Template for Zero-shot CoT in Law Domain.}
\label{fig:legal-prompt-base}
\end{figure*}

\begin{figure*}[h]
\centering
\includegraphics[width=0.95\linewidth]{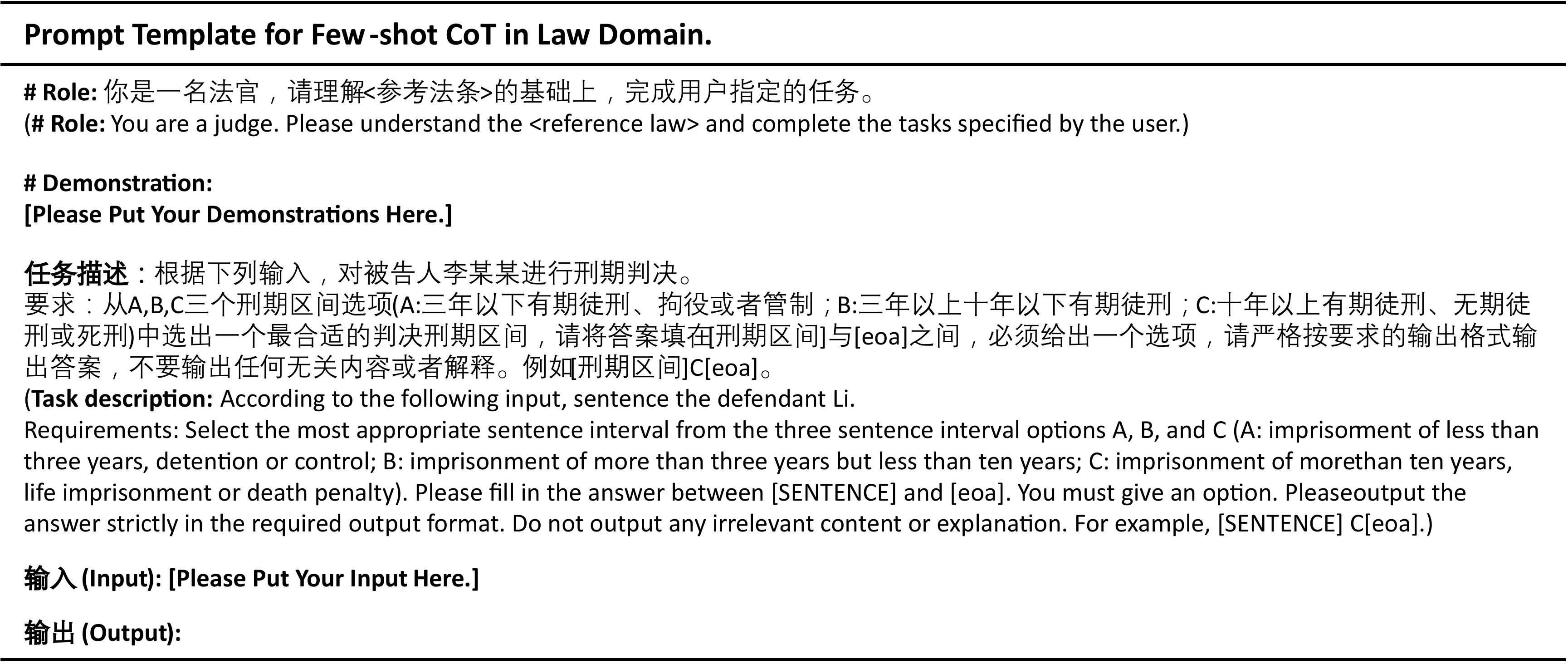}
\caption{The Prompt Template for Few-shot CoT in Law Domain.}
\label{fig:legal-prompt-few}
\end{figure*}

\begin{figure*}[h]
\centering
\includegraphics[width=0.95\linewidth]{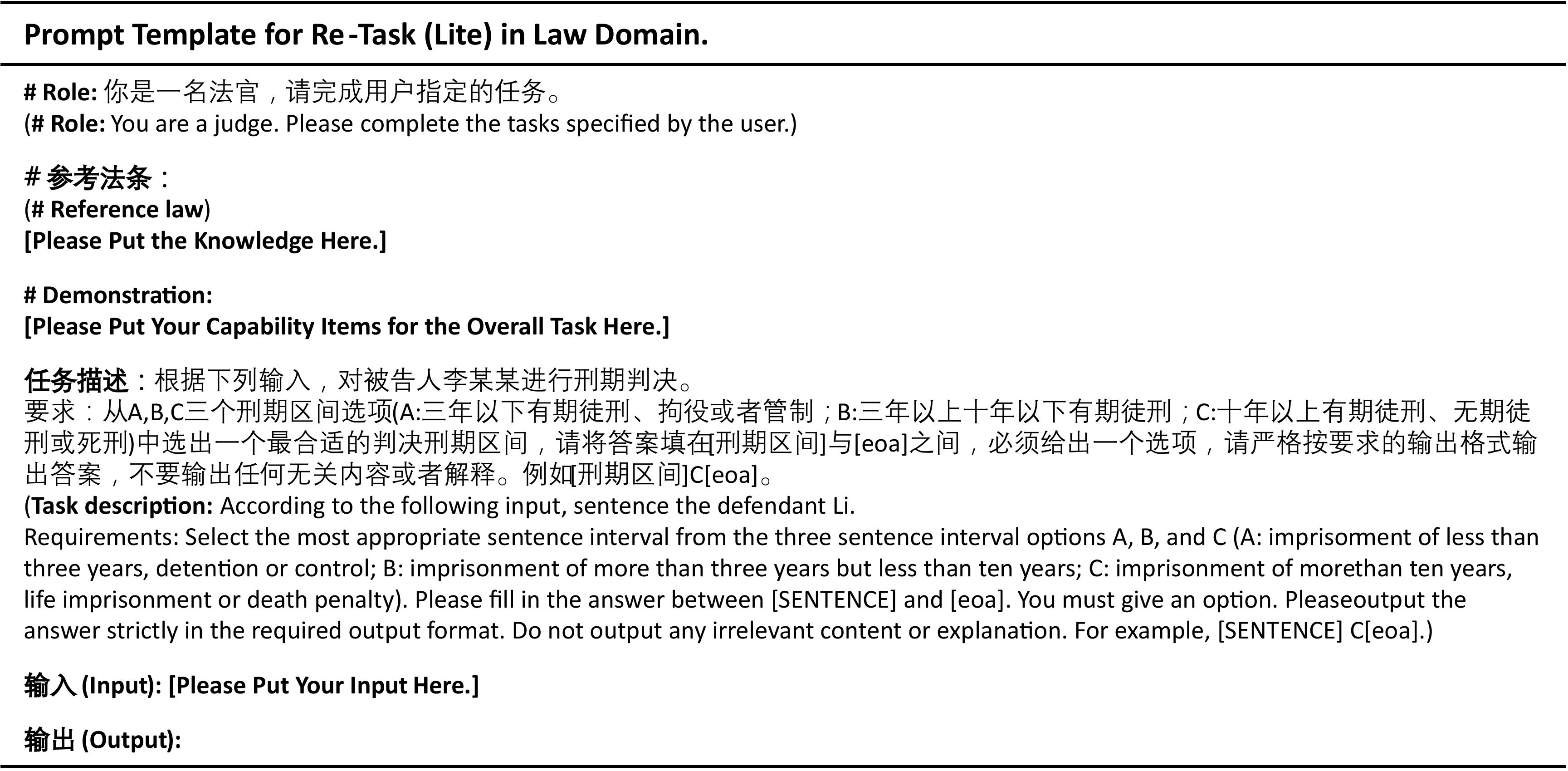}
\caption{The Prompt Template for Re-TASK (Lite) in Law Domain.}
\label{fig:legal-prompt-retask}
\end{figure*}

The prompt templates using zero-shot CoT strategy, few-shot CoT strategy and Re-TASK (Lite) strategy for the sentencing prediction task in the law domain are shown in Figure~\ref{fig:legal-prompt-base}, Figure~\ref{fig:legal-prompt-few}, Figure~\ref{fig:legal-prompt-retask}.

\subsection{STEM Domain}
The prompt templates using zero-shot CoT strategy, few-shot CoT strategy and Re-TASK (Lite) strategy for the MMLU datasets in the STEM domains are shown in Table~\ref{tab:math_template_base}, Table~\ref{tab:math_template_demo} and Table~\ref{tab:math_template_retask}.

\input{Tables/math_template_base}

\input{Tables/math_template_demo}
\input{Tables/math_template_retask}

The prompt templates for generating knowledge and capability items for the MMLU datasets in the STEM domains are shown in Table~\ref{tab:math_template_generatek0} and Table~\ref{tab:math_template_democot}.

\input{Tables/math_template_generatek0}
\input{Tables/math_template_democot}

\subsection{Financial Domain}
\begin{figure*}[h]
\centering
\includegraphics[width=\linewidth]{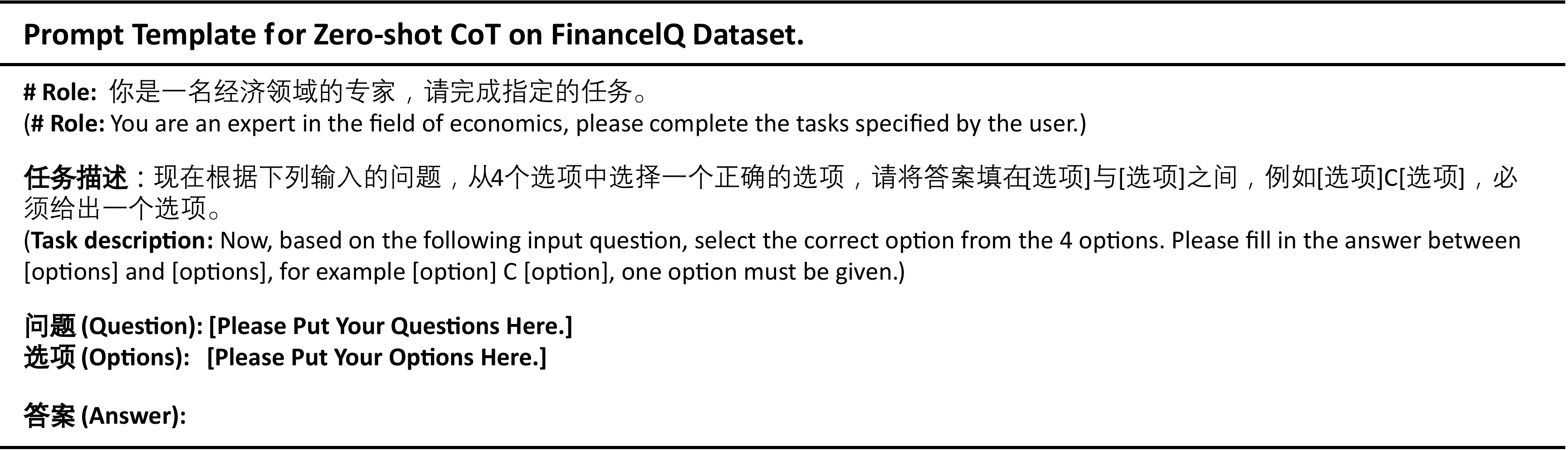}
\caption{The Prompt Template for Zero-shot CoT on FinanceIQ Dataset.}
\label{fig:financial-prompt-base}
\end{figure*}

\begin{figure*}[h]
\centering
\includegraphics[width=\linewidth]{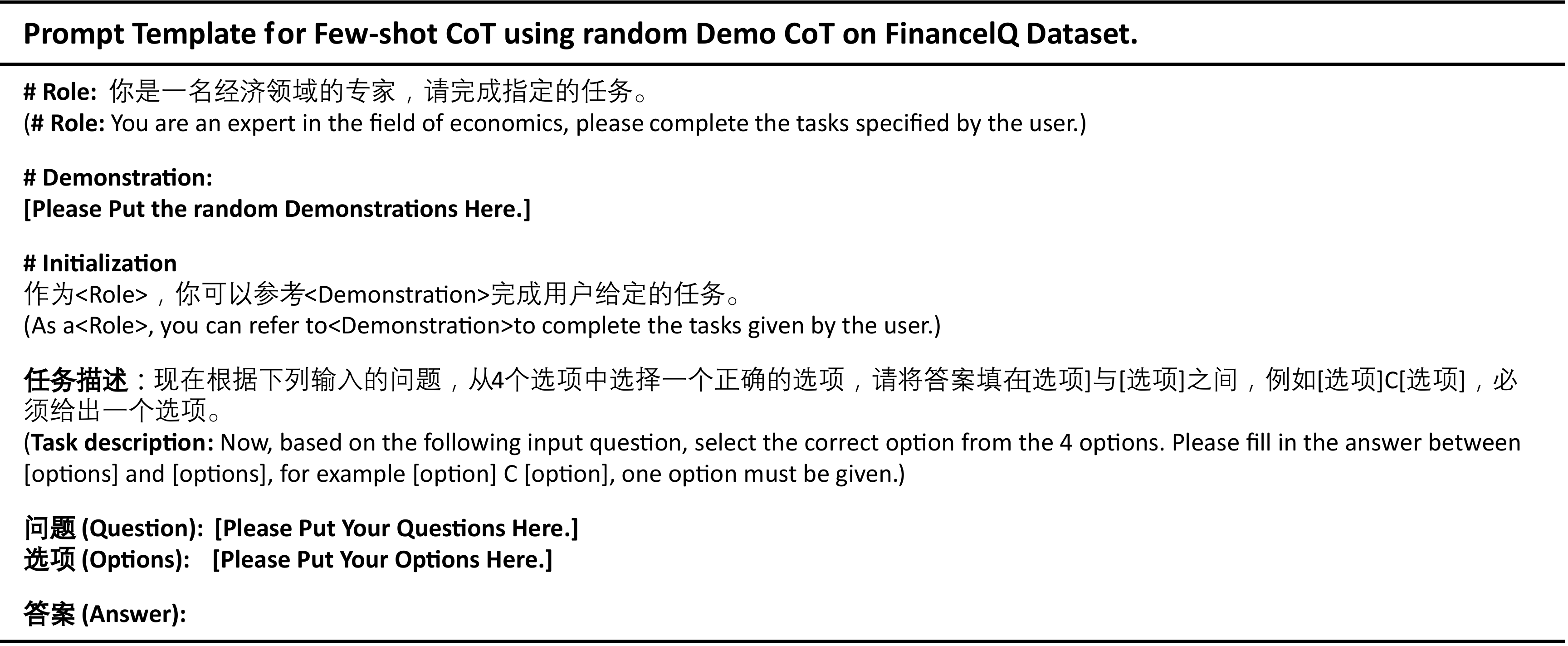}
\caption{The Prompt Template for Few-shot CoT on FinanceIQ Dataset.}
\label{fig:financial-prompt-few}
\end{figure*}

\begin{figure*}[h]
\centering
\includegraphics[width=\linewidth]{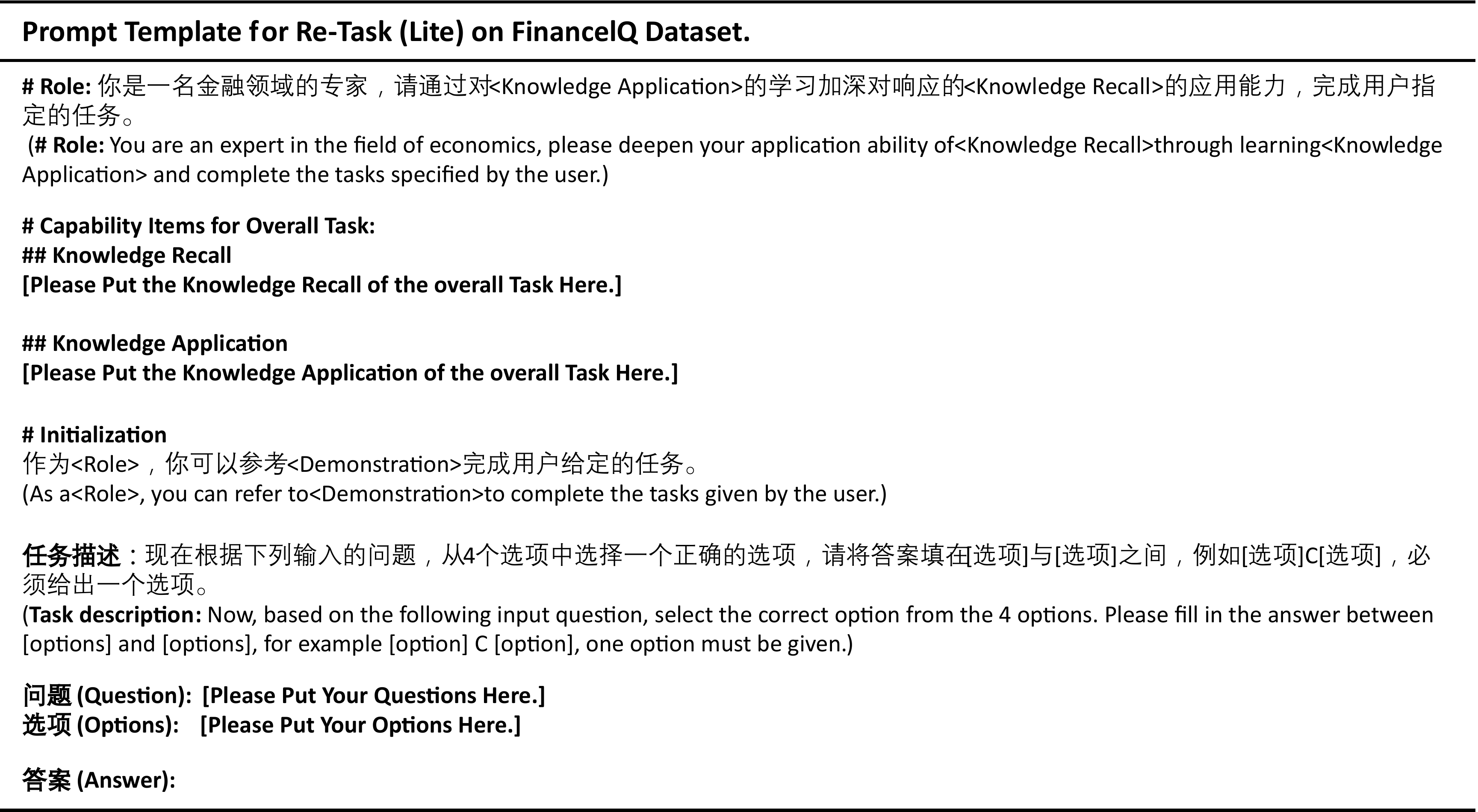}
\caption{The Prompt Template for Re-TASK (Lite) on FinanceIQ Dataset.}
\label{fig:financial-prompt-retask}
\end{figure*}

\begin{figure*}[h]
\centering
\includegraphics[width=\linewidth]{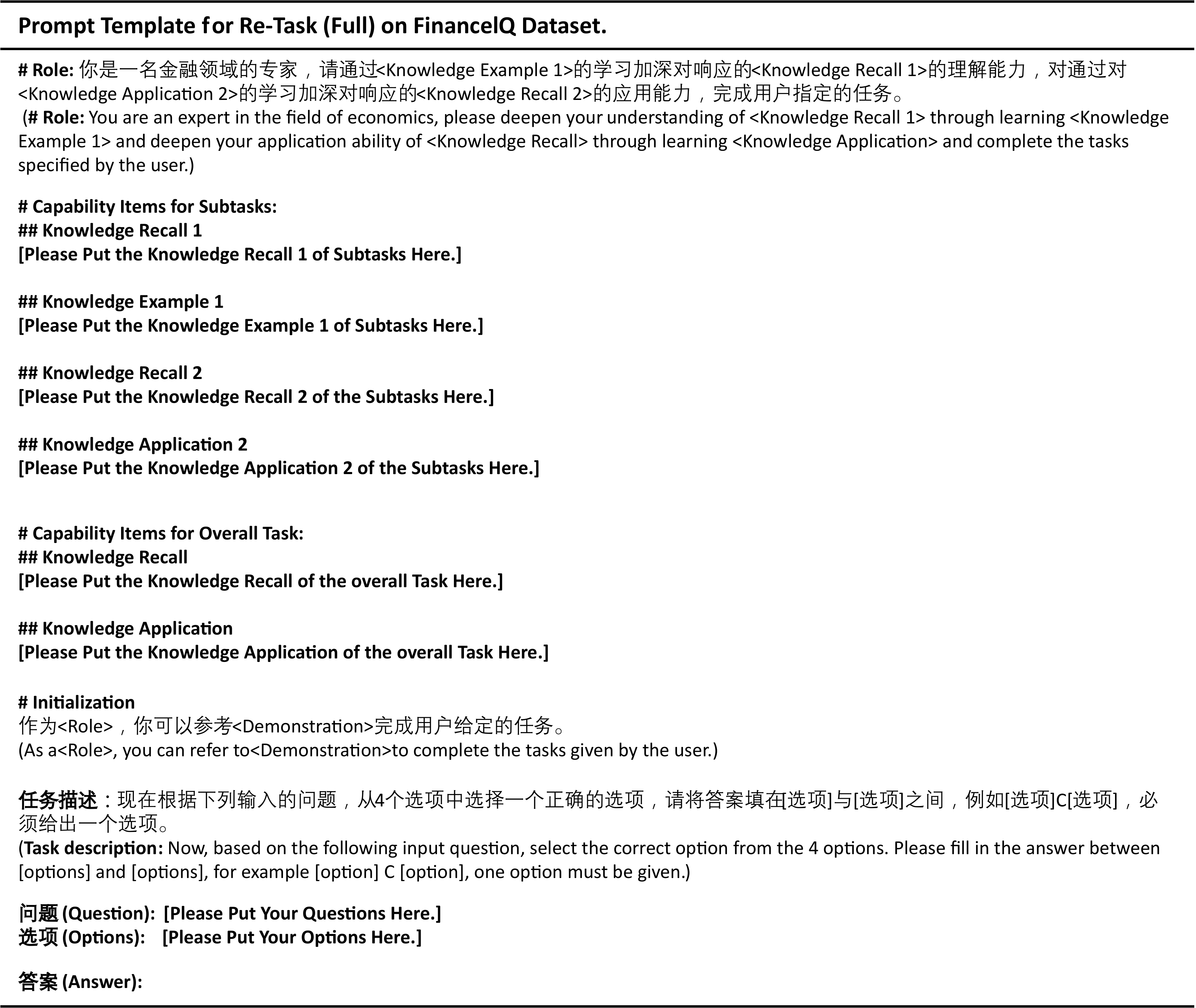}
\caption{The Prompt Template for Re-TASK (Full) on FinanceIQ Dataset.}
\label{fig:financial-prompt-retaskfull}
\end{figure*}

The prompt templates using zero-shot CoT strategy, few-shot CoT strategy, Re-TASK (Lite) strategy and Re-TASK (Full) strategy in the financial domain are shown in Figure~\ref{fig:financial-prompt-base}, Figure~\ref{fig:financial-prompt-few}, Figure~\ref{fig:financial-prompt-retask} and Figure~\ref{fig:financial-prompt-retaskfull}.

\begin{figure*}[h]
\centering
\includegraphics[width=\linewidth]{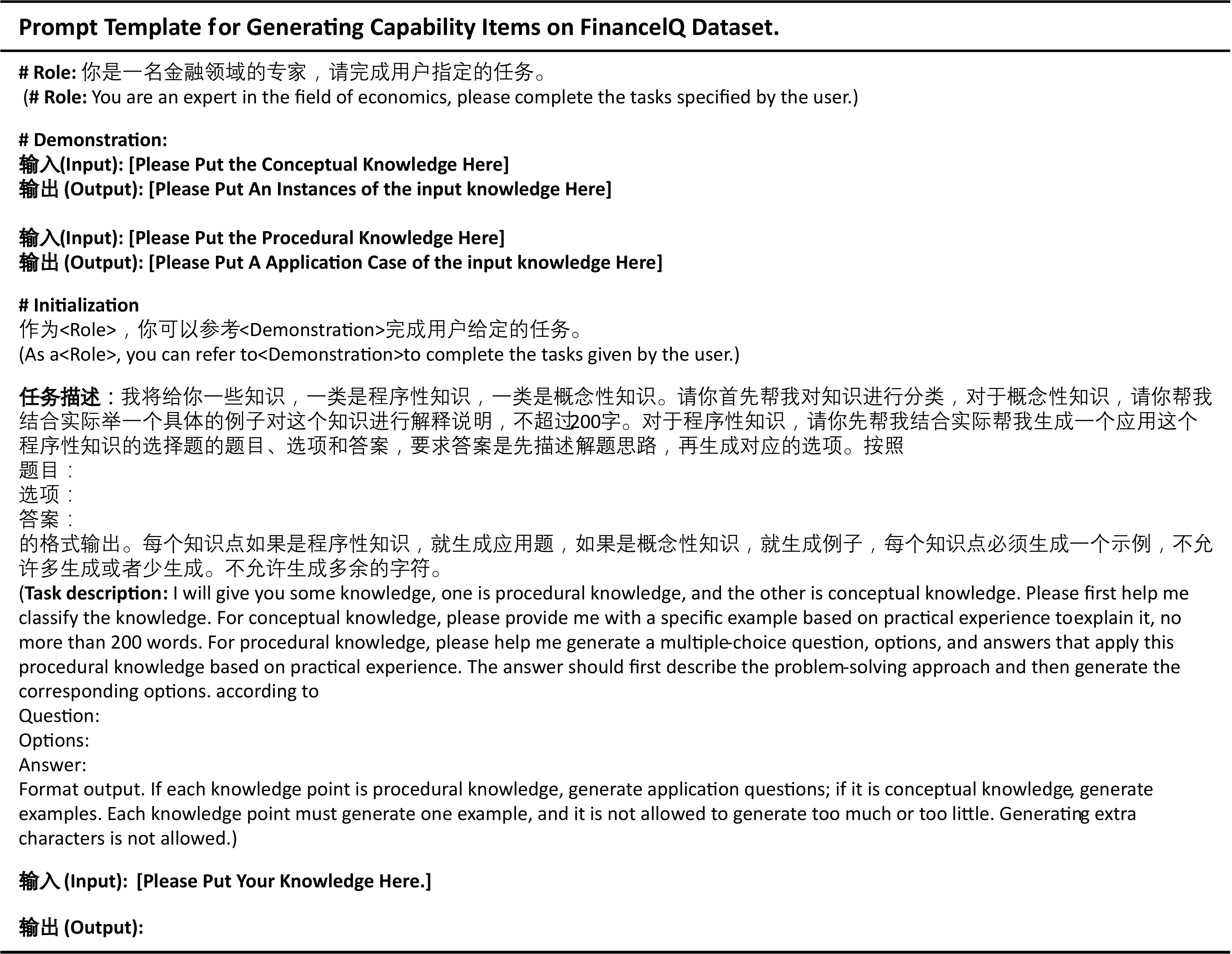}
\caption{The Prompt Template for generating capability items on FinanceIQ Dataset.}
\label{fig:financial-prompt-cap}
\end{figure*}

The prompt templates for capability items in the financial domains are shown in Figure~\ref{fig:financial-prompt-cap}.


\section{Results of K0 and CAP}
\input{new_experiments/results_in_legal_domain}
\input{new_experiments/results_in_financial_domain}

\end{document}

%% file: Tables/legal_results_new.tex
\begin{table*}[!ht]
\small
\centering
\caption{Comparison of accuracy (\%) across Zero-shot CoT, Few-shot CoT, and Re-TASK strategies in the law domain. “Zero-shot CoT + SC” refers to Zero-shot CoT with self-consistency, and “n-shot CoT” refers to Few-shot CoT with $n$ randomly selected demonstrations.} 
\label{tab:legal-results}
\begin{tabularx}{0.95\textwidth}{>{\centering\arraybackslash}p{1.3cm}
    >{\centering\arraybackslash}p{3cm}
    >{\centering\arraybackslash}p{2.8cm}
    >{\centering\arraybackslash}p{1.9cm}
    >{\centering\arraybackslash}p{1.9cm}
    >{\centering\arraybackslash}p{1.9cm}}
\toprule
 \textbf{Methods}     &       \textbf{Prompting Strategies}                    & \textbf{Llama3-Chinese-8B} & \textbf{Yi-1.5-9B}          & \textbf{Qwen1.5-7B}    & \textbf{Average Gain}     \\ \midrule
 
\multirow{6}{*}{\begin{tabular}[c]{@{}c@{}}Traditional\\ CoT\end{tabular}} & Zero-shot CoT         & 54.00                & 40.00             & 33.50           & -                \\

& Zero-shot CoT + SC    & 54.50              & 40.50           & 33.50           & +0.33           \\

& One-shot CoT        & 53.67             & 66.50           & 36.17          & +9.61          \\

& Three-shot CoT        & 56.33             & 70.17          & 38.50           & +12.50         \\ 
&
Plan-and-Solve        & 54.50          & 33.50           & 45.00          & +1.83          \\

& Step-Back        & 72.50             & 72.50          & 36.50           & +18.00         \\ \midrule

\multirow{2}{*}{Re-TASK}   & Re-TASK (+K0)      & 60.50       & 57.50    & 44.00 & +11.50 \\
 & Re-TASK (Lite)      & \textbf{78.50}       & \textbf{85.00}    & \textbf{45.50} & \textbf{+27.17} \\
 \bottomrule
\end{tabularx}
\vspace{2pt}
\end{table*}

%% file: Tables/legal_efficiency.tex
\begin{table*}[]
\small
\centering
\caption{Comparison of token length acorss various prompting strategies in the law domain.}
\begin{tabularx}{0.95\textwidth}{>{\centering\arraybackslash}p{1.6cm}
    >{\centering\arraybackslash}p{3.6cm}
    >{\centering\arraybackslash}p{3cm}
    >{\centering\arraybackslash}p{2.4cm}
    >{\centering\arraybackslash}p{2.4cm}}
\toprule
\textbf{Methods}     &       \textbf{Prompting Strategies} & \textbf{Llama3-Chinese-8B} & \textbf{Yi-1.5-9B} & \textbf{Qwen1.5-7B} \\ \midrule
\multirow{3}{*}{\begin{tabular}[c]{@{}c@{}}Traditional\\CoT\end{tabular}}  &Zero-shot CoT & 526 & 510 & 431  \\
&One-shot CoT & 1691 & 1185 & 1007  \\
&Three-shot CoT & 3104 & 2292 & 2176 \\ \midrule

\multirow{1}{*}{Re-TASK}&Re-TASK (Lite) & {1291} & {1071} & {967}  \\
 \bottomrule
\end{tabularx}
\label{table:tokens}
\vspace{1pt}
\end{table*}

%% file: Tables/financial_results_new.tex
\begin{table*}[ht]
\small
\centering
\caption{Comparison of accuracy (\%) across different prompt strategies on the FinancelQ dataset.}\label{tab:financialresults}
\begin{tabularx}{0.95\textwidth}{>{\centering\arraybackslash}p{1.3cm}
    >{\centering\arraybackslash}p{3cm}
    >{\centering\arraybackslash}p{2.8cm}
    >{\centering\arraybackslash}p{1.9cm}
    >{\centering\arraybackslash}p{1.9cm}
    >{\centering\arraybackslash}p{1.9cm}}
\toprule
\textbf{Methods}& \textbf{Prompting Strategies} & \textbf{Llama3-Chinese-8B} & \textbf{Yi-1.5-9B}          & \textbf{Qwen1.5-7B}     & \textbf{Average Gain}  \\ \midrule
  
\multirow{6}{*}{\begin{tabular}[c]{@{}c@{}}Traditional\\ CoT\end{tabular}} &Zero-shot CoT                    & 36.52          & 53.93          & 43.82          & -                   \\

&Zero-shot CoT + SC             & 34.27          & 61.80          & 46.63          & +2.81                \\
 
&One-shot CoT             & 34.69          & 64.33          & 46.07          & +3.60                \\
  
&Three-shot CoT               & 34.27          & 63.82          & 46.07          & +3.30                \\ 
& Plan-and-Solve        & 30.34             & 66.29           & 41.01          & +1.12          \\

& Step-Back        & 30.90             & 66.85          & 44.38           & +2.62        \\ \midrule

\multirow{3}{*}{Re-TASK} &Re-TASK (+K0)  & 34.27 & 66.29 & 39.89 & +2.06       \\

&Re-TASK (Lite)  & 38.20 & 61.80 & 49.44 & +5.06       \\

&Re-TASK (Full)  & \textbf{52.81} & \textbf{73.60} & \textbf{51.69} & \textbf{+14.61}      \\ \bottomrule

\end{tabularx}
\vspace{2pt}
\end{table*}

%% file: Tables/STEM.tex
\begin{table*}[ht]
\small
  \centering
  \caption{Comparison accuracy (\%) across different prompt strategies on STEM datasets.}
    \begin{tabularx}{0.95\textwidth}{>{\centering\arraybackslash}p{1.5cm}
    >{\centering\arraybackslash}p{3.2cm}
    >{\centering\arraybackslash}p{2cm}
    >{\centering\arraybackslash}p{2cm}
    >{\centering\arraybackslash}p{2cm}
    >{\centering\arraybackslash}p{2cm}}
    \toprule
 \textbf{Domains}    &   \textbf{Prompting Strategies}  & \textbf{Llama3-8B} &  \textbf{Mistral-7B} & \textbf{Qwen1.5-7B} &\textbf{Average Gain} \\
    \midrule
 \multirow{6}{*}{Math} &   Zero-shot CoT & 40.58    & 24.28 & 36.96 & - \\
  &  Zero-shot CoT+SC & 48.19    & 24.64 & 41.67 & +4.23  \\
  &  One-shot CoT & 49.42    & 23.41 & 36.52 & +2.51  \\

 &   Plan-and-Solve & 40.58    & 30.43 & 23.91 & -2.30 \\
  &  Step-Back & 45.65    & \textbf{34.42} & 19.93 & -0.60  \\
    \cmidrule(lr){2-6}
&  Re-TASK (Lite) & \textbf{51.81}    & {28.99} & \textbf{43.84} & \textbf{+7.61}  \\
    \midrule

 \multirow{6}{*}{Physics} &   Zero-shot CoT & 57.84    & 37.25 & 42.16 & - \\
  &  Zero-shot CoT+SC & 58.82    & 39.22 & 37.25 & -0.65  \\
  &  One-shot CoT & \textbf{60.78}    & \textbf{45.10} & 42.16 & +3.59  \\

    &  Plan-and-Solve & 55.88  & 42.16 & 41.18 & +0.65  \\
  &  Step-Back & 30.39    & 34.31 & 30.39 & -14.05  \\
    \cmidrule(lr){2-6}
&  Re-TASK (Lite) & \textbf{60.78}    & {44.12} & \textbf{50.98} & \textbf{+6.21}  \\
    \midrule

 \multirow{6}{*}{Biology} &   Zero-shot CoT & 76.39    & 57.64 & 59.72 & - \\
  &  Zero-shot CoT+SC & 78.47    & 60.42 & 62.50 & +2.55  \\
  &  One-shot CoT & 79.86    & 68.75 & 60.42 & +5.09  \\
    
    &  Plan-and-Solve & 73.61    & 55.56 & 61.11 & -1.16  \\
  &  Step-Back & 43.75    & 50.69 & 53.47 & -15.28  \\
    \cmidrule(lr){2-6}
&  Re-TASK (Lite) & \textbf{88.19}    & \textbf{79.17} & \textbf{81.25} & \textbf{+18.29}  \\
    \bottomrule
    \end{tabularx}%
  \label{tab:stem}%
\vspace{1pt}
\end{table*}%

%% file: Tables/legal_experimental_results_model_scale.tex
\begin{table*}[h]
\centering
\small
\caption{Accuracy comparison (\%) across zero-shot CoT and Re-TASK (Lite) using various scales of Qwen1.5 models on the sentencing prediction dataset.} \label{Table:modelscale}
\begin{tabularx}{0.75\textwidth}{
    >{\centering\arraybackslash}p{3.5cm}
    >{\centering\arraybackslash}p{2.3cm}
    >{\centering\arraybackslash}p{2.3cm}
    >{\centering\arraybackslash}p{2.3cm}}
\toprule
\textbf{Prompting Strategies}   & \textbf{Qwen1.5-7B} & \textbf{Qwen1.5-14B} & \textbf{Qwen1.5-32B} \\ \midrule
 Zero-shot CoT & 33.50 & 48.50 & 84.00 \\
 \midrule
Re-TASK (Lite) & \textbf{45.5} & \textbf{81.17} & \textbf{88.33} \\
 \bottomrule
\end{tabularx}
\end{table*}

%% file: Tables/model_scale_fin.tex
\begin{table*}[ht]
\centering
\small
\caption{Accuracy comparison (\%) across different prompt strategies using various scales of Qwen1.5 models on FinancelQ dataset.} \label{Table:modelscale_fin}
\begin{tabularx}{0.75\textwidth}{
    >{\centering\arraybackslash}p{3.5cm}
    >{\centering\arraybackslash}p{2.3cm}
    >{\centering\arraybackslash}p{2.3cm}
    >{\centering\arraybackslash}p{2.3cm}}
\toprule
\textbf{Prompting Strategies}   & \textbf{Qwen1.5-7B} & \textbf{Qwen1.5-14B} & \textbf{Qwen1.5-32B} \\ \midrule
 Zero-shot CoT & 43.82 & 49.19 & 60.67 \\
 \midrule
Re-TASK (Lite) & 49.44 & 57.30 & 61.80 \\
Re-TASK (Full) & \textbf{51.69} & \textbf{57.87} & \textbf{71.91}\\
 \bottomrule
\end{tabularx}
\end{table*}

%% file: Tables/financial_length.tex
\begin{table*}[ht]
\centering
\small
\caption{Comparison of token length across various prompting strategies on FinanceIQ dataset}
\begin{tabularx}{0.8\textwidth}{
    >{\centering\arraybackslash}p{3.3cm}
    >{\centering\arraybackslash}p{3.2cm}
    >{\centering\arraybackslash}p{2.2cm}
    >{\centering\arraybackslash}p{2.2cm}}
\toprule
\textbf{Prompting Strategies} & \textbf{Llama3-Chinese-8B} & \textbf{Yi-9B} & \textbf{Qwen1.5-7B} \\
 \midrule
Zero-shot CoT & 861 & 541 & 417 \\
One-shot CoT & 1138 & 844 & 767 \\
Three-shot CoT & 1730 & 1478 & 1421 \\
\midrule
Re-TASK (Lite)& 1032 &742 & 602 \\
Re-TASK (Full) & \textbf{1713} & \textbf{1707} & \textbf{1581}\\
\bottomrule
\end{tabularx}
\label{tab:financial_length}
\end{table*}

%% file: Tables/math_token.tex
\begin{table*}[ht]
  \centering
  \small
  \caption{Comparison of token length across various prompting strategies on MMLU-Math dataset.}
\begin{tabularx}{0.8\textwidth}{
    >{\centering\arraybackslash}p{3.3cm}
    >{\centering\arraybackslash}p{3.2cm}
    >{\centering\arraybackslash}p{2.2cm}
    >{\centering\arraybackslash}p{2.2cm}}
    \toprule
      \textbf{Prompting Strategies}     & \textbf{Llama3-8B} & \textbf{Qwen1.5-7B} & \textbf{Mistral-7B} \\
    \midrule
    Zero-shot CoT & 301  & 449  & 544  \\
    One-shot CoT & 615  & 615  & 768  \\
    \midrule
    Re-TASK(Lite) & \textbf{628}  & \textbf{673}  & \textbf{769}  \\
    \bottomrule
    \end{tabularx}%
  \label{tab:math_token}%
\end{table*}%

%% file: Tables/math_case1_base.tex
\begin{table*}[h]
  \centering
  \caption{An example of the Prompt and the Response for Zero-shot CoT strategy on MMLU-Math dataset.}
    \begin{tabular}{p{\textwidth}}
    \toprule
    \textbf{Prompt and Response for Standard Zero-shot CoT} \\
    \midrule
    \textbf{Input:} \\
    \# Role: \\
    You are an expert in the field of Math. Complete the task provided by the user. \\\\
    \# Task Description: \\
    According to the following question and options, select one correct option. Please think step by step and follow the answer format, i.e., output the brief chain-of-thought beginning with 'Rationale: ' within 500 characters, and the final choice option beginning with 'Correct: ' from 'A' to 'D'. \\
    Question: \\
    The length of a rectangle is twice its width. Given the length of the diagonal is \$5\textbackslash{}sqrt{5}\$, find the area of the rectangle. \\
    Options:  \\
    A. 2500; B. 2; C. 50; D. 25 \\
     \\
    \textbf{Output:} \\
    Rationale: Let the width of the rectangle be x. Since the length is twice the width, the length is 2x.\\
    The diagonal is the hypotenuse of a right triangle with legs x and 2x. Using the Pythagorean theorem, we can write: \\
    x\^{}2 + (2x)\^{}2 = (5\textbackslash{}sqrt{5})\^{}2 \\
    Simplifying the equation, we get: \\
    5x\^{}2 = 25 \\
    Dividing both sides by 5, we get: \\
    x\^{}2 = 5 \\
    Taking the square root of both sides, we get: \\
    x = $\sqrt{5}$ \\
    Now, the area of the rectangle is: \\
    Area = length × width = 2x × x = 2($\sqrt{5}$)\^{}2 = 2(5) = 10 \\
    Correct: D. 25 \\
    \bottomrule
    \end{tabular}%
  \label{tab:math_case1_base}%
\end{table*}%

%% file: Tables/math_case1_retask.tex
\begin{table*}[h]
  \centering
  \caption{An example of the Prompt and the Response for Re-TASK (Lite) strategy on MMLU-Math dataset.}
    \begin{tabular}{p{\textwidth}}
    \toprule
    \textbf{Prompt and Response for Re-TASK Prompting} \\
    \midrule
    \textbf{Input:} \\
    \# Role:  \\
    You are an expert in the field of Math. Complete the task provided by the user. \\
     \\
    \# Knowledge:  \\
    The area of a rectangle is given by A = length * width, and the Pythagorean theorem states that for a right triangle, a\^{}2 + b\^{}2 = c\^{}2, where c is the length of the hypotenuse (diagonal). \\\\
    \# Demonstration:  \\
    Question: In a rectangle, the length is 5 cm and the width is 12 cm. What is the length of the diagonal? \\
    Options:  \\
    A. 13 cm; B. 14 cm; C. 15 cm; D. 16 cm \\
    Rationale:  \\
    Step 1. We're given a rectangle with length 5 cm and width 12 cm. Step 2. We can use the Pythagorean theorem to find the length of the diagonal. Step 3. The diagonal is the hypotenuse of a right triangle with legs 5 cm and 12 cm. Step 4. Using the theorem, we get diagonal\^{}2 = 5\^{}2 + 12\^{}2 = 25 + 144 = 169. Step 5. Taking the square root, we get diagonal = 13 cm. The correct answer is A. \\
    Correct: A \\
     \\
    \# Task Description:  \\
    According to the following question and options, refer to $<$\# Knowledge$>$ and $<$\# Demonstration$>$, select one correct option. Please think step by step and follow the answer format in $<$\# Demonstration$>$, i.e., output the brief chain-of-thought beginning with 'Rationale: ' within 500 characters, and the final choice option beginning with 'Correct: ' from 'A' to 'D'.  \\
    Question:  \\
    The length of a rectangle is twice its width. Given the length of the diagonal is \$5\textbackslash{}sqrt{5}\$, find the area of the rectangle. \\
    Options:  \\
    A. 2500; B. 2; C. 50; D. 25 \\
     \\
     \textbf{Output:}\\
    Rationale:  \\
     \\
    Step 1. We're given a rectangle with length = 2 * width. Step 2. We're also given the length of the diagonal, which is \$5\textbackslash{}sqrt{5}\$. Step 3. We can use the Pythagorean theorem to find the length and width. Step 4. Let the width be x, then the length is 2x. Step 5. Using the theorem, we get (2x)\^{}2 + x\^{}2 = (\$5\textbackslash{}sqrt{5}\$)\^{}2. Step 6. Simplifying, we get 4x\^{}2 + x\^{}2 = 125. Step 7. Combining like terms, we get 5x\^{}2 = 125. Step 8. Dividing both sides by 5, we get x\^{}2 = 25. Step 9. Taking the square root, we get x = 5. Step 10. Now we can find the area, which is length * width = 2x * x = 2 * 5 * 5 = 50. \\
    Correct: C \\
    \bottomrule
    \end{tabular}%
  \label{tab:math_case1_retask}%
\end{table*}%

%% file: Tables/modelsource.tex
\begin{table*}[h]
\centering
\small
\renewcommand{\arraystretch}{1.5}
\caption{Models, Sources and Licenses Used in This Work}
\begin{tabular}{clc}
\toprule
\textbf{Models} & \textbf{Model sources}                                                                                                   & \textbf{License}   \\ \midrule

Qwen1.5-7B      & https://huggingface.co/Qwen/Qwen1.5-7B-Chat                                                                              & Apache License 2.0 \\

Qwen1.5-14B     & https://huggingface.co/Qwen/Qwen1.5-14B-Chat                                                                             & Apache License 2.0 \\

Qwen1.5-32B     & https://huggingface.co/Qwen/Qwen1.5-32B-Chat                                                                             & Apache License 2.0 \\

Yi-1.5-9B        & https://huggingface.co/01-ai/Yi-1.5-9B-Chat                                                                              & Apache License 2.0 \\

Llama3-Chinese-8B       & \renewcommand{\arraystretch}{1.1}\begin{tabular}[c]{@{}l@{}}https://www.modelscope.cn/models/FlagAlpha/\\ Llama3-Chinese-8B-Instruct/summary\end{tabular}
& Apache License 2.0 \\ 

Llama3-8B                       & \renewcommand{\arraystretch}{1.1}\begin{tabular}[c]{@{}l@{}}https://www.modelscope.cn/models/FlagAlpha/\\ Llama3-8B-Instruct/summary\end{tabular} & Apache License 2.0\\
Mistral-7B      & https://huggingface.co/mistralai/Mistral-7B-Instruct-v0.2    & Apache License 2.0 \\ 
Llama3.1-70B                    & https://huggingface.co/meta-llama/Meta-Llama-3.1-70B-Instruct                 & llama3.1 license           \\
Qwen2.5-72B & https://huggingface.co/Qwen/Qwen2.5-72B-Instruct & Qwen license\\\bottomrule

\end{tabular}

\label{tab:modelsource}
\end{table*}

%% file: Tables/datasource.tex
\begin{table*}[h]
\centering
\small
\caption{Datasets and sources used in this work}
\label{tab:datasources}
\renewcommand{\arraystretch}{1.6}
\begin{tabular}{lll}
\toprule
\textbf{Datasets} & \textbf{Sources}                                    \\ \midrule

MMLU              & https://huggingface.co/datasets/cais/mmlu                 \\
FinanceIQ & https://huggingface.co/datasets/Duxiaoman-DI/FinanceIQ \\
CAIL 2018 & https://github.com/thunlp/CAIL?tab=readme-ov-file 
\\ \bottomrule
\end{tabular}
\end{table*}

%% file: Tables/math_template_base.tex
\begin{table*}[h]
  \centering
  \caption{The prompt template of standard zero-shot CoT on MMLU-Math dataset.}
    \begin{tabular}{p{\textwidth}}
    \toprule
    \textbf{Prompt template of Standard Zero-shot CoT} \\
    \midrule
    \# Role: \\
    You are an expert in the field of Math. Complete the task provided by the user. \\\\
    \# Task Description: \\
    According to the following question and options, select one correct option. Please think step by step and follow the answer format, i.e., output the brief chain-of-thought beginning with 'Rationale: ' within 500 characters, and the final choice option beginning with 'Correct: ' from 'A' to 'D'. \\
    Question: \\
    $[$Please Put Your Question Here$]$\\
    Options:  \\
    $[$Please Put Your Options Here$]$ \\
    \bottomrule
    \end{tabular}%
  \label{tab:math_template_base}%
\end{table*}%

%% file: Tables/math_template_demo.tex
\begin{table*}[h]
  \centering
  \caption{The prompt template of one-shot CoT on MMLU-Math dataset.}
    \begin{tabular}{p{\textwidth}}
    \toprule
    \textbf{Prompt template of One-shot CoT} \\
    \midrule
    \textbf{Input:} \\
    \# Role:  \\
    You are an expert in the field of Math. Complete the task provided by the user. \\
     \\
    \# Demonstration:  \\
    Question: \\
    $[$Please Put Your Question of Demonstration Here$]$ \\
    Options:  \\
    $[$Please Put Your Options of Demonstration Here$]$ \\
    Rationale:  \\
    $[$Please Put Your Rationale of Demonstration Here$]$\\
    Correct: $[$Please Put Your Final Choice of Demonstration Here$]$ \\
     \\
    \# Task Description:  \\
    According to the following question and options, refer to $<$\# Demonstration$>$, select one correct option. Please think step by step and follow the answer format in $<$\# Demonstration$>$, i.e., output the brief chain-of-thought beginning with 'Rationale: ' within 500 characters, and the final choice option beginning with 'Correct: ' from 'A' to 'D'.  \\
    Question:  \\
    $[$Please Put Your Question Here$]$\\
    Options:  \\
    $[$Please Put Your Options Here$]$ \\
    \bottomrule
    \end{tabular}%
  \label{tab:math_template_demo}%
\end{table*}%

%% file: Tables/math_template_retask.tex
\begin{table*}[htbp]
  \centering
  \caption{The prompt template of Re-TASK prompt on MMLU-Math dataset.}
    \begin{tabular}{p{\textwidth}}
    \toprule
    \textbf{Prompt template of Re-TASK} \\
    \midrule
    \textbf{Input:} \\
    \# Role:  \\
    You are an expert in the field of Math. Complete the task provided by the user. \\
     \\
    \# Knowledge:  \\
    $[$Please Put Your Knowledge Here$]$\\\\
    \# Demonstration:  \\
    Question: \\
    $[$Please Put Your Question of Demonstration Here$]$ \\
    Options:  \\
    $[$Please Put Your Options of Demonstration Here$]$ \\
    Rationale:  \\
    $[$Please Put Your Rationale of Demonstration Here$]$\\
    Correct: $[$Please Put Your Final Choice of Demonstration Here$]$ \\
     \\
    \# Task Description:  \\
    According to the following question and options, refer to $<$\# Knowledge$>$ and $<$\# Demonstration$>$, select one correct option. Please think step by step and follow the answer format in $<$\# Demonstration$>$, i.e., output the brief chain-of-thought beginning with 'Rationale: ' within 500 characters, and the final choice option beginning with 'Correct: ' from 'A' to 'D'.  \\
    Question:  \\
    $[$Please Put Your Question Here$]$\\
    Options:  \\
    $[$Please Put Your Options Here$]$ \\
    \bottomrule
    \end{tabular}%
  \label{tab:math_template_retask}%
\end{table*}%

%% file: Tables/math_template_generatek0.tex
\begin{table*}[htbp]
  \centering
  \caption{The prompt template of knowledge generation.}
    \begin{tabular}{p{\textwidth}}
    \toprule
    \textbf{Prompt template of Knowledge Generation} \\
    \midrule
    \# Role: \\
    You are an expert in the field of Math. Complete the task provided by the user. \\\\
    \# Demonstration:  \\
    \#\# Question: The hypotenuse of a right triangle measures 10 inches and one angle is \$45\^{}\{\textbackslash circ\}\$. What is the number of square inches in the area of the triangle?\\
    \#\# Knowledge: The area of a right triangle is given by A = (1/2) * base * height.\\\\
    \# Task Description: \\
    Given the question, please just generate the formula or other knowledge related to the question as brief as possible, like the $<$\# Demonstration$>$. Just output the one related formula or other knowledge, DO NOT output any other characters.\\
    \#\# Question: \\
    $[$Please Put Your Question Here$]$\\
    \#\# Knowledge:  \\
    \bottomrule
    \end{tabular}%
  \label{tab:math_template_generatek0}%
\end{table*}%


%% file: Tables/math_template_democot.tex
\begin{table*}[htbp]
  \centering
  \caption{The prompt template of capability item generation.}
    \begin{tabular}{p{\textwidth}}
    \toprule
    \textbf{Prompt template of Capability Item Generation} \\
    \midrule
    \# Role: \\
    You are an expert in the field of Math. Complete the task provided by the user. \\\\
    \# Demonstration:  \\
    \{\\
    \quad``question":\\
    \quad\quad``At a certain factory, 10 percent of the staplers produced on Monday were defective and 2 percent of the non-defective staplers were rejected by mistake. If 72 of the non-defective staplers were rejected, what was the number of staplers produced that day?" \\
    \quad``options": $[$\\
    \quad\quad``A. 4,000",\\
    \quad\quad``B. 4,200",\\
    \quad\quad``C. 4,500",\\
    \quad\quad``D. 4,800"\\
    \quad$]$\\
    \quad``rationale": \\
    \quad\quad``Step 1. We're told that 10\% of staplers in a factory are defective. \textbackslash n Step 2. X = Total staplers, 0.1X = defective staplers, 0.9X = normal staplers. \textbackslash n Step 3. We're told that 2\% of the normal staplers were rejected by mistake and that this = 72 staplers. \textbackslash n Step 4. 0.9X(0.02) = 72, 0.018X = 72, 18X = 72,000, X = 4,000.", \\
    \quad``correct": ``A"\\
    \}\\\\
    \# Task Description: \\
    I will give you a piece of knowledge text, please help me generate a four-choice question which is one deduction application of this knowledge, including the question, options, rationale and correct answer. \\
    The knowledge is $[$Please Put Your Knowledge Here$]$.\\
    The answer is required to follow the **json** format in $<$\# Demonstration$>$, as:\\\\
    \{\\
    \quad``question": Content of the question,\\
    \quad``options": list of four options,\\
    \quad``rationale": Content of the chain-of-thought with each step starting with 'Step x. ', which is limited to 400 characters,\\
    \quad``correct": the single choice character of correct answer\\
    \}\\
    You must and can only generate one deduction example of the given knowledge in the above json format. No extra characters are allowed.\\
    \bottomrule
    \end{tabular}%
  \label{tab:math_template_democot}%
\end{table*}%

%% file: new_experiments/results_in_legal_domain.tex
\begin{table*}[!ht]
\small
\centering
\caption{Comparison of accuracy (\%) across Zero-shot CoT, Few-shot CoT, and Re-TASK strategies in the law domain. “Zero-shot CoT + SC” refers to Zero-shot CoT with self-consistency, and “n-shot CoT” refers to Few-shot CoT with $n$ randomly selected demonstrations.} 
\label{tab:legal-resultsnew}
\vspace{-2pt}
\begin{tabularx}{0.95\textwidth}{>{\centering\arraybackslash}p{1.3cm}
    >{\centering\arraybackslash}p{3cm}
    >{\centering\arraybackslash}p{2.8cm}
    >{\centering\arraybackslash}p{1.9cm}
    >{\centering\arraybackslash}p{1.9cm}
    >{\centering\arraybackslash}p{1.9cm}}
\toprule
\textbf{Methods}        &     \textbf{Prompting Strategies}                      & \textbf{Llama3-Chinese-8B} & \textbf{Yi-1.5-9B}          & \textbf{Qwen1.5-7B}    & \textbf{Average Gain}     \\ \midrule
 
\multirow{6}{*}{\begin{tabular}[c]{@{}c@{}}Traditional\\ CoT\end{tabular}} & Zero-shot CoT         & 54.00                & 40.00             & 33.50           & -                \\

& Zero-shot CoT + SC    & 54.50              & 40.50           & 33.50           & +0.33           \\

& One-shot CoT        & 53.67             & 66.50           & 36.17          & +9.61          \\

& Three-shot CoT        & 56.33             & 70.17          & 38.50           & +12.50         \\ 
&
Plan-and-Solve        & 54.50          & 33.50           & 45.00          & +1.83          \\

& Step-Back        & 72.50             & 72.50          & 36.50           & +18.00         \\ \midrule

\multirow{3}{*}{Re-TASK}   & Re-TASK (+K0)      & 60.50       & 57.50   & 44.00 & +11.50 \\
& Re-TASK (+Cap)      & 49.17       & 55.00    & 34.33 & +3.67 \\
& Re-TASK (Lite)      & \textbf{78.50}       & \textbf{85.00}    & \textbf{45.50} & \textbf{+27.17} \\
 \bottomrule
\end{tabularx}
\end{table*}

%% file: new_experiments/results_in_financial_domain.tex
\begin{table*}[ht]
\small
\centering
\caption{Comparison of accuracy (\%) across different prompt strategies on FinancelQ dataset.}\label{tab:financialresultsnew}
\begin{tabularx}{0.95\textwidth}{>{\centering\arraybackslash}p{1.3cm}
    >{\centering\arraybackslash}p{3cm}
    >{\centering\arraybackslash}p{2.8cm}
    >{\centering\arraybackslash}p{1.9cm}
    >{\centering\arraybackslash}p{1.9cm}
    >{\centering\arraybackslash}p{1.9cm}}
\toprule
\textbf{Methods}        &     \textbf{Prompting Strategies}     & \textbf{Llama3-Chinese-8B} & \textbf{Yi-1.5-9B}          & \textbf{Qwen1.5-7B}     & \textbf{Average Gain}  \\ \midrule
  
\multirow{6}{*}{\begin{tabular}[c]{@{}c@{}}Traditional\\ CoT\end{tabular}} &Zero-shot CoT                    & 36.52          & 53.93          & 43.82          & -                   \\

&Zero-shot CoT + SC             & 34.27          & 61.80          & 46.63          & +2.81                \\
 
&One-shot CoT             & 34.69          & 64.33          & 46.07          & +3.60                \\
  
&Three-shot CoT               & 34.27          & 63.82          & 46.07          & +3.30                \\ 
& Plan-and-Solve        & 30.34             & 66.29           & 41.01          & +1.12          \\

& Step-Back        & 30.90             & 66.85          & 44.38           & +2.62        \\ \midrule

\multirow{4}{*}{Re-TASK} &Re-TASK (K0)  & 34.27 & 66.29 & 39.89  & +2.06     \\
&Re-TASK (+Cap)  & 33.71 & 62.36 & 38.76 & +0.19       \\
&Re-TASK (Lite)  & 38.20 & 61.80 & 49.44 & +5.06       \\
&Re-TASK (Full)  & \textbf{52.81} & \textbf{73.60} & \textbf{51.69} & \textbf{+14.61}      \\ \bottomrule

\end{tabularx}
\end{table*}